# Compass-v3: Scaling Domain-Specific LLMs for Multilingual E-Commerce in Southeast Asia

**Sophia Maria**[1]
[1]Shopee LLM Team

Large language models (LLMs) excel in general-domain applications, yet their performance often degrades in specialized tasks requiring domain-specific knowledge. E-commerce is particularly challenging, as its data are noisy, heterogeneous, multilingual, and highly dynamic. We present Compass-v3, a vertical-domain Mixture-of-Experts (MoE) model with 245B total parameters and 71B active per token, designed for Southeast Asian e-commerce. Compass-v3 adopts fewer but larger experts, combined with hardware-efficient optimizations—such as intra-node expert parallelism and a customized memcpy operator—to maximize GPU utilization. The model is trained on 12T tokens of curated multilingual corpora and large-scale synthetic e-commerce instructions using a mixed-training strategy. To enhance alignment, we propose Optimal-Transport Direct Preference Optimization (OTPO), which captures token-level distinctions and improves instruction adherence in commerce-specific scenarios. Extensive evaluations demonstrate that Compass-v3 delivers state-of-the-art e-commerce performance, surpassing DeepSeek-V3.1, GPT-4 series, and Qwen3-235B. Moreover, Compass-v3 demonstrates strong multilingual capability across low-resource Southeast Asian languages (Indonesian, Thai, Filipino, Vietnamese, Malay, Taglog) and Portuguese while sustaining competitive performance on general benchmarks. It has already been widely applied in Shopee's industrial-scale e-commerce platform and is gradually replacing OpenAI's traffic, now accounting for over 70% of total LLM usage, highlighting its dual strengths in specialized commerce expertise and broad linguistic competence.

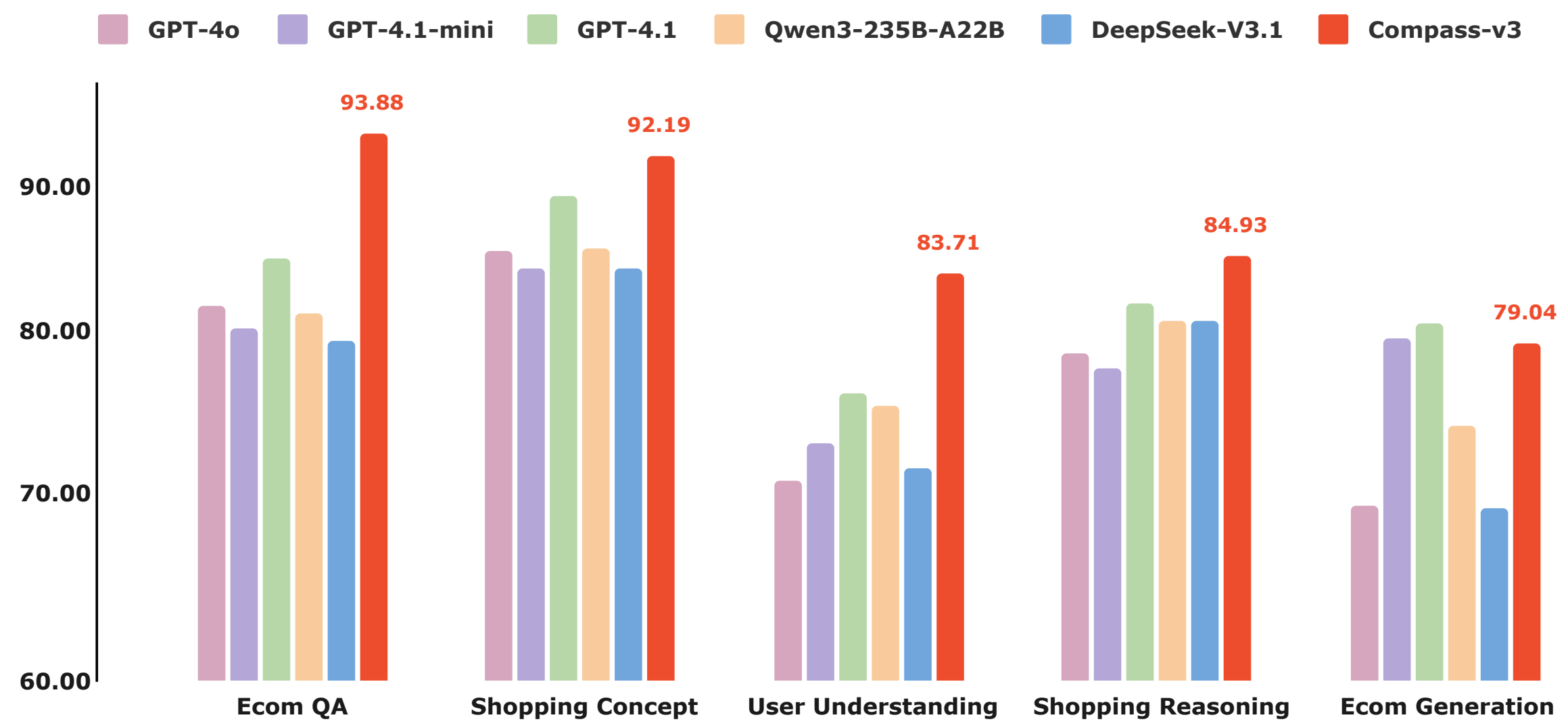

Figure 1 | Compass-v3 demonstrates superior performance over competitors on e-commerce tasks.

*Corresponding author(s): sophia.maria@shopee.com*

## 1. Introduction

Large language models (LLMs) have achieved remarkable progress in general-purpose language understanding and generation. Models like GPT-4 (Achiam et al., 2023)) reach impressive performance across many tasks, yet even the best general LLMs often struggle in specialized domains requiring in-depth knowledge or domain-specific skills (Peng et al., 2024). This limitation has sparked a rise of domain-specific LLMs targeting particular fields such as programming, finance, and medicine. For example, in the coding domain OpenAI's Codex (Chen et al., 2021) achieved 28.8% solve rate on a Python code benchmark where a general GPT-3 solved 0%. In finance, the BloombergGPT (Wu et al., 2023) model trained on financial corpora outperforms general models on financial tasks without sacrificing general NLP performance. Likewise in medicine, Med-PaLM (Tu et al., 2024) was the first to exceed the physician passing score on USMLE (Dillon et al., 2004) medical exam questions, demonstrating the value of aligning LLMs to domain expertise. These successes illustrate that while general LLMs are powerful, domain-specific LLMs can substantially bridge the gap in specialized settings by incorporating domain-focused data and knowledge.

The e-commerce domain presents unique and challenging terrain for LLMs. E-commerce data is notoriously noisy and heterogeneous: user queries and reviews contain typos, slang, and code-mixed language; product descriptions and specifications have semi-structured formats; and content spans multiple modalities and styles (Ren et al., 2024). Moreover, e-commerce platforms must handle multilingual interactions - especially in Southeast Asia (SEA), where many languages are low-resource - while adapting to highly dynamic content with constantly emerging products, trends, and user behaviors. General-purpose LLMs, even at large scales, often struggle with these demands due to limited exposure to the specific jargon and rapidly evolving e-commerce context. To address this, recent works have introduced e-commerce–specific LLMs (Geng et al., 2022; Li et al., 2024; Peng et al., 2024; Shi et al., 2023), such as LLaMA-E (Shi et al., 2023), fine-tuned for e-commerce content generation (e.g. ad text, product Q&A), and EcomGPT (Li et al., 2024), an instruction-tuned model leveraging chain-of-task e-commerce data. However, these early efforts have limitations. Some only cover a narrow subset of tasks (Shi et al., 2023), and others heavily rely on synthetic or repurposed instruction data with limited realism (Li et al., 2024). Furthermore, most existing LLMs have primarily focused on English and Chinese, leaving a significant gap in support for low-resource SEA languages. **In summary, the e-commerce domain requires a dedicated LLM that can tackle noisy, multi-format, multilingual, and dynamic content – needs not fully met by prior general models or early domain-specific attempts.**

To this end, we present **Compass-v3, a vertical-domain large language model tailored for SEA e-commerce**. This is made possible through **a combination of architectural, optimization, and alignment innovations.** Architecturally, the model adopts a sparse Mixture-of-Experts (MoE) architecture with fewer but substantially larger experts, each with high activation capacity, thereby delivering strong representational power under limited compute budgets. To further mitigate resource constraints, we design *training and inference optimizations*. For training, we apply *intra-node expert parallelism* over NVLink and a *memcpy operator optimization* that removes redundant copies in MoE module. For inference, we apply *expert-aware FP8 quantization* which doubles decoding speed while preserving accuracy across low-resource languages. Central to its performance is a *domain-centric data strategy*: we mine large volumes of high-quality e-commerce text and construct extensive *synthetic instruction datasets* that cover search, recommendation, customer service, and compliance workflows. A mixed-training regime integrate both general-purpose and vertical-domain corpora, striking a balance between robustness and specialized competence. For alignment, we design *Optimal-Transport Direct Preference Optimization (OTPO) (Li et al., 2025)*, which captures fine-grained, token-level distinctions in preference pairs, thereby strengthening instruction adherence in complex e-commerce scenarios. Collectively, these design choices enable Compass-v3 to deliver state-of-the-art accuracy in e-commerce tasks while retaining competitive general performance.

Through extensive experimental comparisons, **Compass-v3 demonstrates state-of-the-art performance across e-commerce, multilingual, and general benchmarks.** In e-commerce tasks, Compass-v3 achieves leading results on both in-house and open-source benchmarks. It excels in practical business scenarios such as product guidance, after-sales service, and product understanding, outperforming both proprietary and open-source baselines. Beyond domain specialization, Compass-v3 shows robust multilingual capability, delivering consistently high accuracy across SEA languages including Indonesian, Malay, Thai, Vietnamese, and Portuguese, while maintaining stability across low-resource settings. Importantly, these domain-specific strengths do not come at the cost of general ability: Compass-v3 remains competitive on standard English benchmarks, often approaching GPT-4–mini-class models in reading comprehension, commonsense reasoning, and multilingual understanding. This balanced performance profile underscores Compass-v3's dual advantage of specialized e-commerce expertise and broad general competence, making it well-suited for real-world deployment in diverse SEA markets.

## 2. Pretraining

In this part, we present a resource-efficient framework for building vertical-domain large language models, exemplified by **Compass-v3**, a 245B-parameter Mixture-of-Experts (MoE) model with 71B active parameters, specialized for e-commerce and Southeast Asian (SEA) multilingual tasks.

### 2.1. Architecture

Our model adopts a large-scale Mixture-of-Experts (MoE) architecture, comprising 245B total parameters with 71B active parameters per forward pass. The network integrates 16 large experts, of which 4 experts are dynamically selected by the router for each token. This design enables substantial scaling of model capacity while preserving computational tractability, as illustrated in Figure 2. Our architectural decisions are guided by the following principles:

- **Choice of Large Experts**. Unlike prior works that employ numerous small experts, we deliberately opt for large experts. Small-expert designs often suffer from inefficiencies in Group GEMM operations, where fragmented matrix multiplications significantly reduce hardware utilization. By consolidating capacity into fewer but larger experts, we markedly improve computational efficiency while maintaining the representational richness of MoE models.
- **Multi-Token Prediction (MTP)**. To improve efficiency and predictive capability, we adopt a multi-token prediction (MTP) framework that trains auxiliary layers to predict multiple future tokens in parallel from shifted inputs. Both the main next-token prediction and the auxiliary MTP task share embeddings and the LM head, but parameter updates are strictly decoupled: shared components and the backbone are optimized only by the main loss, while MTP-specific layers are updated solely by the MTP loss. This separation preserves core language modeling quality while enabling MTP heads to exploit frozen representations, thereby stabilizing training, accelerating convergence, and mitigating exposure bias.
- **Rationale for Larger Activation**. We further increase the number of active parameters per step to maximize the effective learning capacity under a limited training budget. By activating more parameters, the model can acquire stronger reasoning and inference capabilities in a shorter training horizon, which is especially valuable in data-constrained regimes. This strategy balances efficiency and performance, yielding faster improvements in downstream reasoning and generation quality.

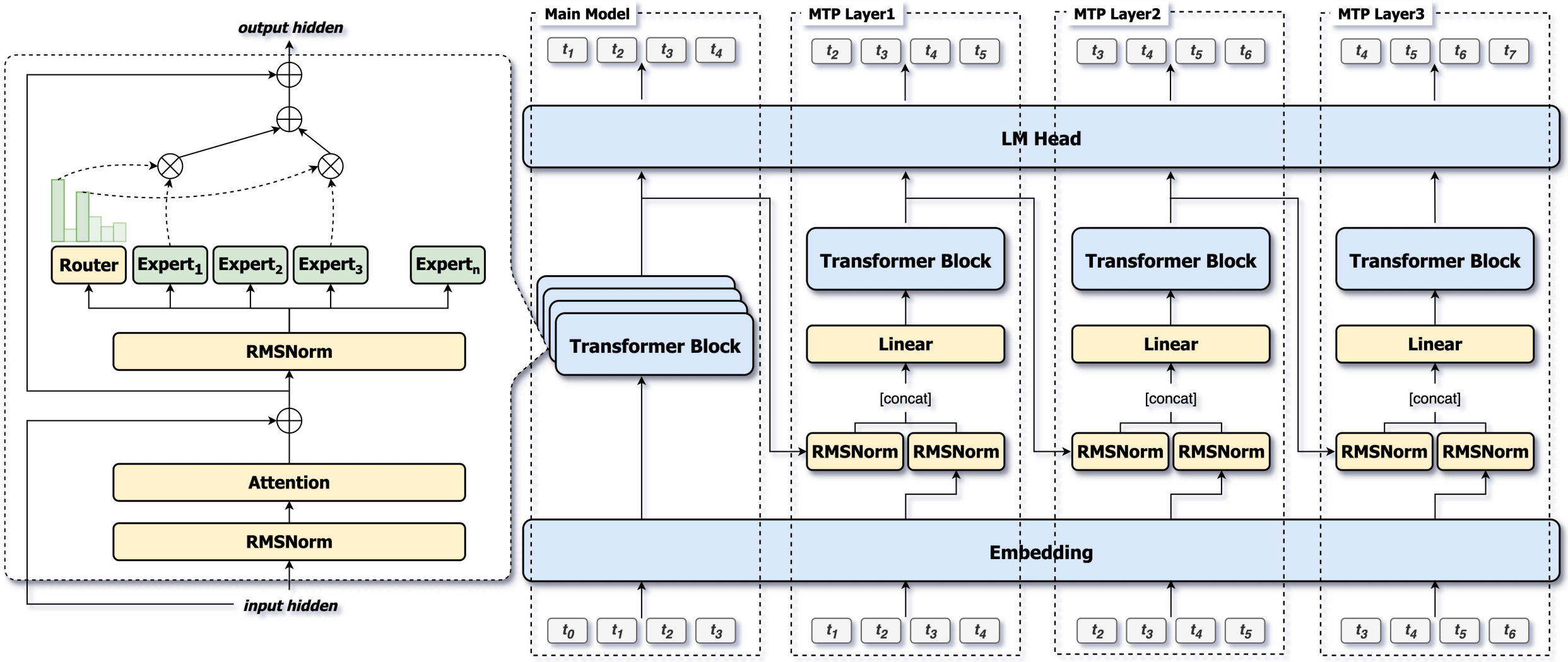

Figure 2 | Overall architecture of our model. The backbone adopts a Transformer enhanced with sparse Mixture-of-Experts (MoE) layers, where a router activates a small subset of experts per token to increase capacity efficiently. On top of the main autoregressive path, we introduce multi-token prediction (MTP) layers that take shifted sequences, normalize and project them through lightweight Transformer blocks, and share a unified embedding and LM head for parallel prediction of multiple future tokens. This design jointly improves efficiency, scalability, and predictive accuracy.

## 2.2. E-commerce-enhanced Training Pipeline

Our training pipeline is designed to strengthen the capabilities of E-commerce and Southeast Asia (SEA) multilingualism, while improving the efficiency of inference through multitoken prediction (MTP) and extending the context length of the model to 128K tokens using long-context techniques. The process unfolds in several progressive stages, as shown in Figure 3:

- **General Knowledge Pretraining**. In the initial stage, the model acquires broad linguistic and reasoning ability. Compared to general-purpose models, we incorporate a significantly higher proportion of e-commerce and SEA multilingual data, thereby maintaining general knowledge while enhancing domain-specific multilingual capacity.
- **E-Commerce and SEA Multilingual Enhancement**. The second stage leverages carefully curated E-commerce tasks and SEA language corpora to continuously strengthen both dimensions, ensuring robustness across low-resource languages and commercial applications.
- **Reasoning Enhancement**. We then emphasize reasoning-focused and high-quality instruction data, improving the model's analytical ability while extending the context window to 8K tokens.
- **Long-Context Extension**. We employ specialized long-context training strategies to extend the model's context window to 128K tokens, enabling robust processing of substantially longer sequences. The extension is performed in two phases: the model is first adapted to 32K tokens with targeted training data, and then further scaled to the full 128K context length.
- **Multi-Token Prediction (MTP)**. Finally, MTP training is interleaved throughout the pipeline, providing faster convergence, more efficient inference, and reduced exposure bias by enabling parallel prediction of multiple future tokens.

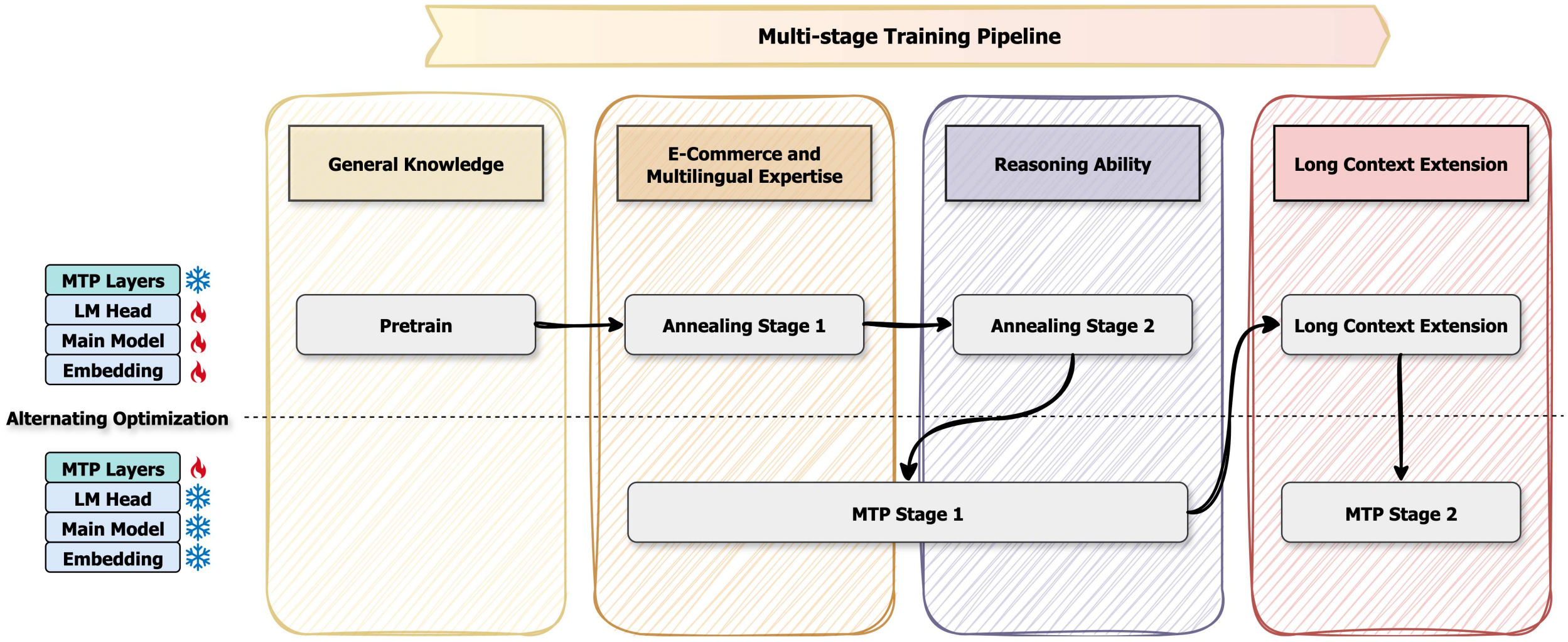

Figure 3 | Multi-stage training pipeline. The model is first pretrained on general knowledge, then annealed toward E-commerce domain expertise and South-East multilingual data, and finally extended to long-context corpora. Multi-token prediction (MTP) is interleaved in two stages, with alternating optimization that updates the backbone during pretraining/annealing while freezing it during MTP training. This design progressively builds general knowledge, domain expertise, reasoning ability, and long-context capability while ensuring stable optimization.

## 2.3. Balanced and Efficient MoE Training

Scaling large language models requires overcoming critical bottlenecks in distributed training systems. We present a suite of system-level optimizations that substantially accelerate pretraining efficiency. Our contributions include (i) intra-node expert parallelism (EP) leveraging NVLink for fast GPU communication, (ii) enhancing computational efficiency under memory constraints via virtual and uneven pipeline parallelism with selective recomputation, (iii) overlapping and fusing all-to-all communication for MoE layers, (iv) memcpy kernel optimization targeting unpermutation hot spots, and (v) auxiliary loss–based expert load balancing. Collectively, these strategies deliver significant throughput improvements while maintaining convergence quality, offering practical guidance for training increasingly large models at scale.

### 2.3.1. *Intra-Node Expert Parallelism*

We configure expert parallelism (EP) within a single node by setting EP=8 and fully exploiting NVLink bandwidth, as shown in Figure 4. Compared to inter-node communication, intra-node EP yields significantly faster throughput and reduces synchronization overhead, thereby accelerating training efficiency without sacrificing flexibility.

### 2.3.2. *Memcpy Kernel Optimization in MoE Layers*

After optimizing intra-node EP communication, we observe a substantial improvement in overall training efficiency. Mixture-of-Experts (MoE) training typically involves four stages: routing, dispatch, computation, and combination. Profiling results (Figure 5) indicate that the combination stage dominates runtime, becoming the primary bottleneck. A deeper analysis reveals that the majority of this overhead arises from memcpy operations in the combination step. To address this issue, we optimize the corresponding operator by eliminating redundant memory copies and reordering data layouts. This optimization yields further acceleration, particularly in memory-bound phases, and significantly enhances end-to-end MoE training efficiency.

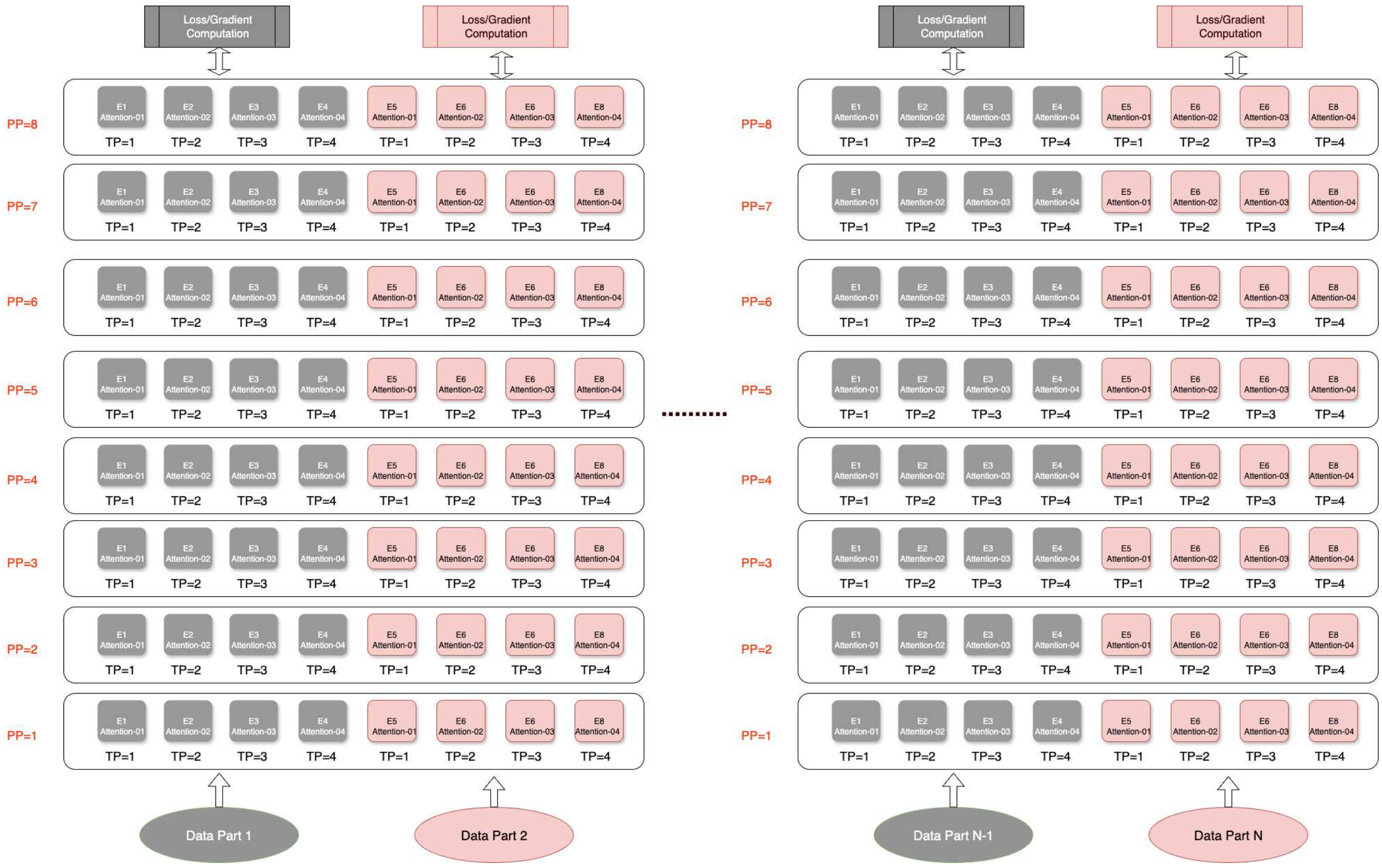

Figure 4 | Our distributed training configuration. Each small square represents a single GPU, with eight GPUs forming one node. We adopt hybrid parallelism by combining data parallelism (DP), pipeline parallelism (PP), tensor parallelism (TP), and intra-node expert parallelism (EP). Expert parallelism is restricted within a node (EP=8), leveraging NVLink for fast intra-node communication and reducing inter-node all-to-all overhead.

### 2.3.3. *Virtual and Uneven Pipeline Parallelism with Selective Recomputation*

Pipeline parallelism helps reduce GPU memory usage, but large pipeline stages inevitably introduce idle bubble time. To mitigate this issue, we first adopt virtual pipeline parallelism (VPP), which increases the number of micro-batches in training, thereby reducing bubble overhead and improving overall efficiency. However, VPP also incurs additional memory consumption, often leading to out-of-memory (OOM) failures. To strike a balance between computational efficiency and memory usage, we apply selective recomputation to the early pipeline stages, where OOM risks are most pronounced.

In addition, we observe that computational loads across pipeline stages are inherently unbalanced, as the first and last stages must perform extra operations such as embedding and loss computation. Selective recomputation further amplifies this imbalance. To address this, we refine the pipeline partitioning strategy by shifting from uniform to uneven layer segmentation, which better aligns the workload distribution across stages and further improves training efficiency.

### 2.3.4. *Expert Load Balancing*

A central challenge in MoE training is balanced token routing, since skewed distributions may underutilize experts or cause router collapse. We address this with two mechanisms. First, we apply

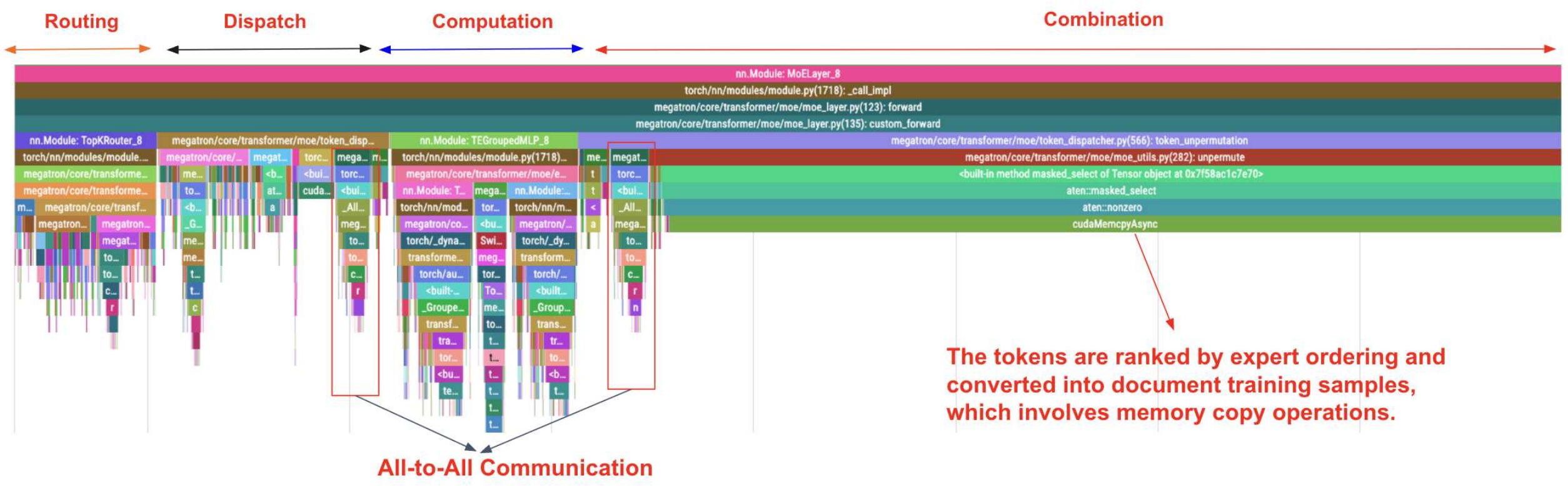

Figure 5 | Profiling of Mixture-of-Experts (MoE) training stages, consisting of routing, dispatch, computation, and combination. Contrary to common assumptions, All-to-All communication is not the dominant bottleneck. Instead, the combination stage consumes the majority of runtime, where repeated memcpy operations in the combination step lead to significant inefficiency.

an auxiliary load-balancing loss.

$$\mathcal{L}_{aux} = N \cdot \sum_{i=1}^{N} \left(\frac{p_i}{B} \cdot \frac{c_i}{B \cdot K}\right)$$

where $N$ is the number of experts, $p_i$ and $c_i$ are the aggregated routing probability and the activation count of expert $i$, respectively. $B$ is the number of tokens in one batch, and $K$ is the number of activated experts for each token.

Second, we stabilize router logits using the Z-loss (Zoph et al., 2022):

$$\mathcal{L}_Z = \frac{1}{B} \sum_{j=1}^{B} (\log \sum_{i=1}^{N} \exp(z_i^j))^2$$

where $z_i^j$ is the routing logits of expert $i$ on token $j$.

To further ensure stability, we disable droptoken, and compute the router's gating layer in fp32 instead of bf16 to avoid overflow. The overall training objective is:

$$\mathcal{L} = \mathcal{L}_{LM} + \alpha \cdot \mathcal{L}_{aux} + \beta \cdot \mathcal{L}_Z$$

where $\alpha$ is the coefficient for auxiliary loss, $\beta$ is the coefficient for Z-loss. Furthermore, to achieve more stable model convergence, both coefficients $\alpha$ and $\beta$ are designed to progressively decay throughout the training process as the number of optimization steps increases.

### *2.3.5. Optimizing Communication Efficiency*

To further enhance communication efficiency, we adopt a series of strategies. Specifically, we implement (i) overlapped gradient reduction which hides synchronization cost behind computation, (ii) overlapped parameter gathering for efficient weight distribution, and (iii) fused MoE All-to-All operations to reduce kernel launch overhead. Collectively, these optimizations substantially reduce end-to-end latency and improve overall GPU utilization.

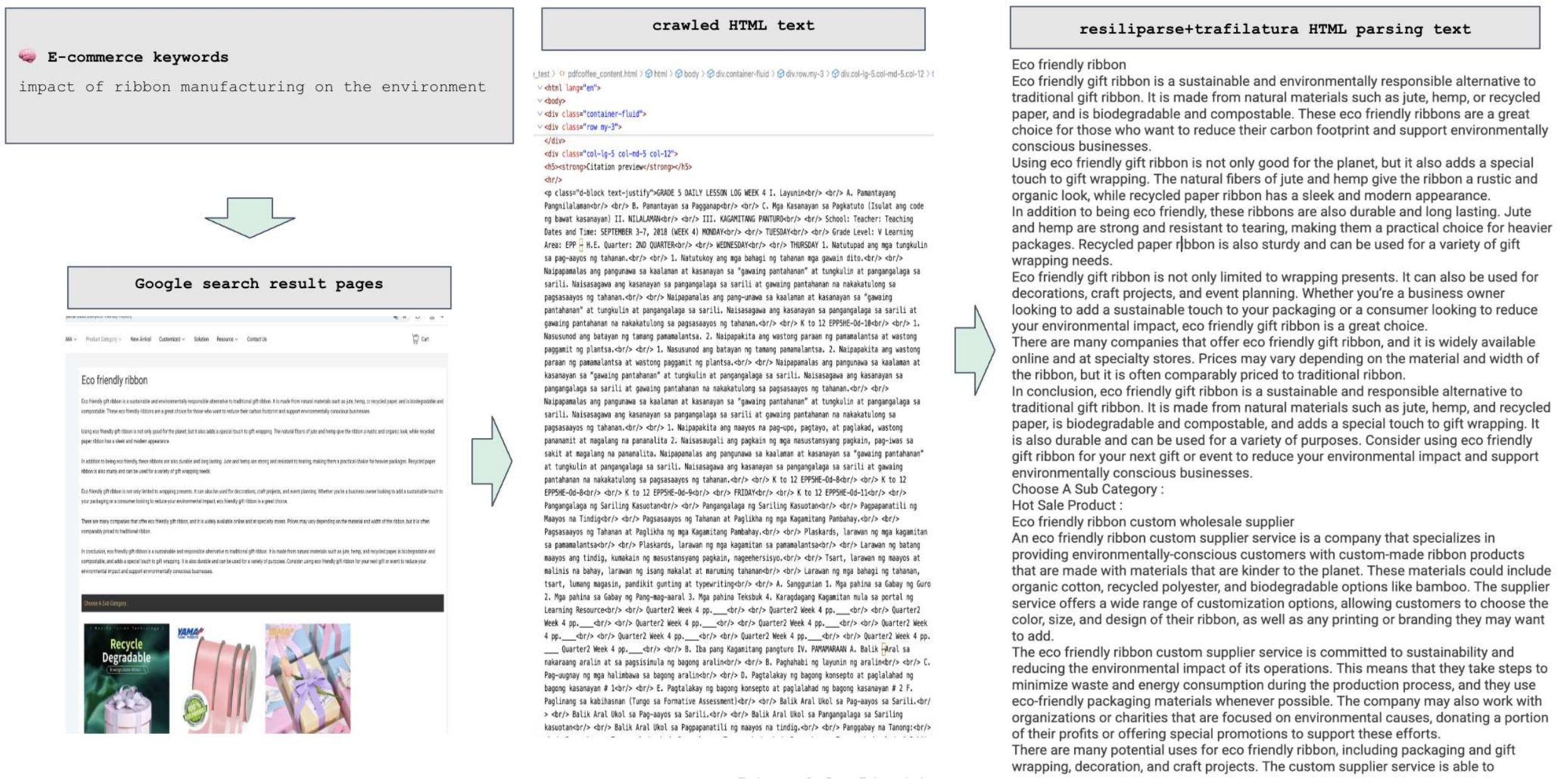

Figure 6 | Keyword-driven web mining pipeline for e-commerce data. Domain-specific queries (e.g., “impact of ribbon manufacturing on the environment”) guide retrieval, followed by HTML crawling and parsing. The resulting corpus captures key e-commerce attributes—Product Descriptions, Usage Scenarios, Consumer Narratives, and Supplier Metadata—yielding a clean, attribute-dense dataset for SEA-focused pretraining.

### 2.4. Data

Training large language models (LLMs) under limited computational resources requires careful curation of high-quality data. We pretrain our model on a 12-trillion-token corpus, carefully constructed to enhance both e-commerce understanding and Southeast Asian (SEA) multilingual capacity. To achieve this, we design a data-centric pipeline that integrates (i) large-scale mining of high-quality e-commerce corpora, (ii) efficient data selection algorithms to maximize information density, and (iii) automated evaluation frameworks for rapid feedback and iteration. This strategy ensures that every token contributes maximally to model utility, enabling us to scale effectively despite compute constraints.

#### *2.4.1. High-Quality E-Commerce Data Mining*

To construct a high-quality corpus tailored for e-commerce applications, we design a keyword-driven web mining pipeline that systematically retrieves, cleans, and structures domain-relevant text. We begin with e-commerce–oriented queries (e.g., “impact of ribbon manufacturing on the environment”) to capture text reflecting product semantics, sustainability discourse, and consumer perspectives. This ensures that the collected data is aligned with commercial intent rather than general web content. Candidate pages are crawled in raw HTML and then processed using robust parsing tools such as resiliparse and trafilatura. The extracted corpus encodes rich commercial attributes, including Product Descriptions, Usage Scenarios, Consumer Narratives, Supplier Metadata. Through this pipeline, we obtain a clean, attribute-dense, and domain-specific corpus that provides comprehensive coverage of e-commerce semantics and SEA-relevant practices, serving as a robust foundation for downstream pretraining.

### 2.4.2. *Efficient Data Selection Algorithms*

Inspired by RegMix (Liu et al., 2025c), we implement a proxy-model-based procedure to search for data mixtures that maximize performance on Southeast Asian languages and e-commerce tasks while preserving strong general capabilities. Following prior mixture-selection practice (Diao et al., 2025; Liu et al., 2025b), we construct the validation set from appropriate benchmarks (e.g., ARC-Easy, ARC-Challenge) by concatenating QA pairs, and run pilot experiments to ensure proxy feedback predicts downstream performance: Across mixtures, proxy validation loss is strongly correlated with benchmark scores, showing that validation loss predicts downstream performance. In addition, the 2M proxy's validation loss is strongly correlated with the 150M proxy's validation loss when trained on the same 1B tokens, establishing cross-scale consistency. Therefore, the 2M proxy's validation loss serves as a reliable predictor of benchmark performance.

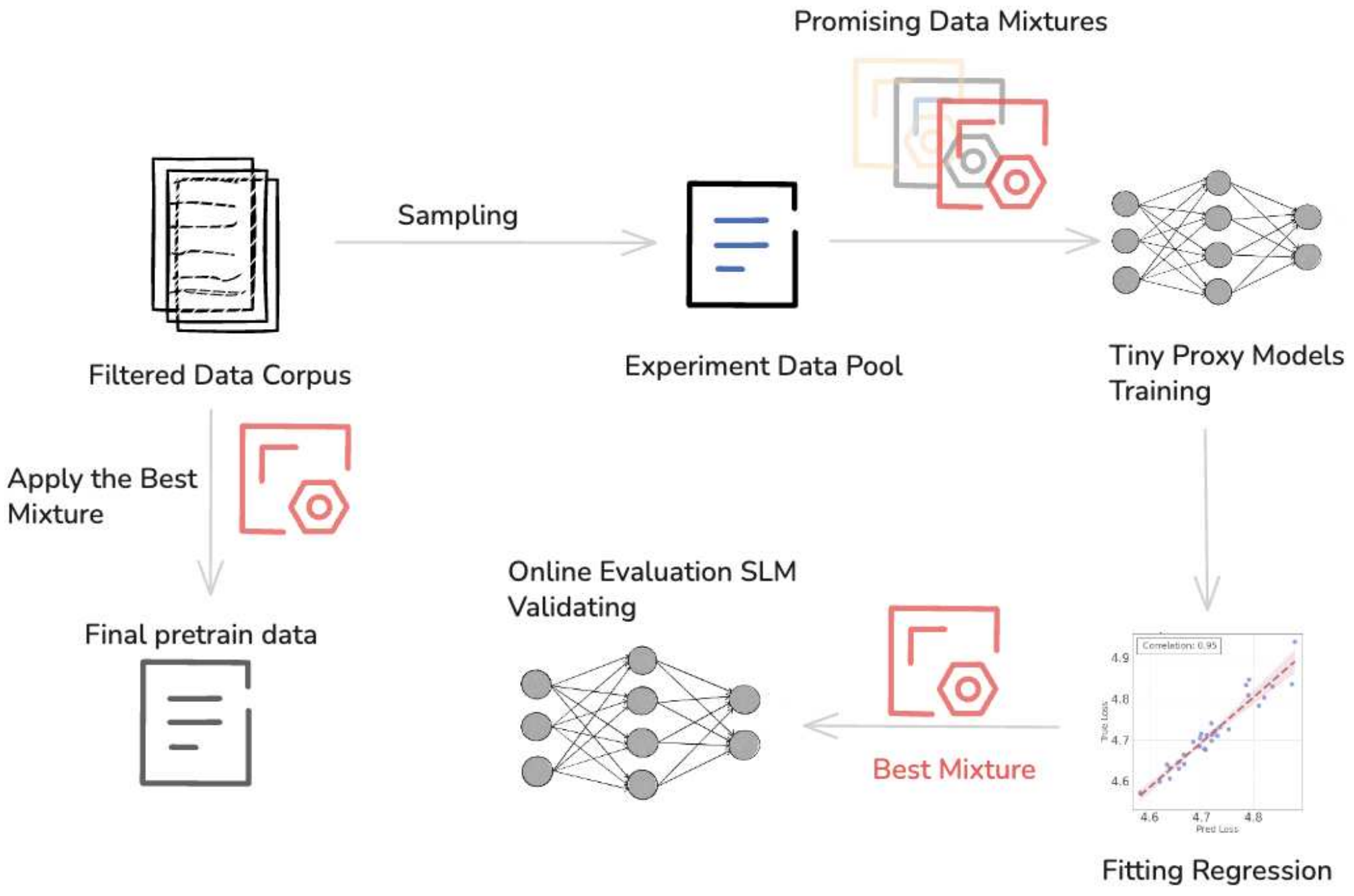

Figure 7 | Data selection pipeline. From the filtered corpus, we build an experiment data pool mirroring the 12T pretraining sampling. We sample 512 mixtures from 16 language/domain shards and train 2M-parameter proxies for 1B tokens using a QA-based validation set (e.g., ARC-Easy, ARC-Challenge) retained after correlation screening. A LightGBM regressor predicts validation loss to pick the best mixture, supported by a 1.3B params with 150B tokens online evaluation; the chosen mixture is then applied to construct the final pretraining data.

We then conduct the mixture search. The filtered corpus is partitioned into 16 language/domain shards, and a curated data pool is created to enforce sampling fidelity: the proxy's 1B tokens are drawn with shard proportions matched to those used for sampling the final 12T-token pretraining corpus (Feng et al., 2024). From these shards we sample 512 diverse candidate mixtures and, for each, train a tiny proxy model (2M non-embedding parameters) for 1B tokens. A LightGBM regressor is fit on mixture descriptors to predict validation loss, and we select the mixture with the lowest predicted loss. Finally, we validate the chosen mixture by training a 1.3B-parameter model for 150B tokens, achieving significant gains on both English and Southeast Asia benchmarks.

#### *2.4.3. Rapid Automated Data Evaluation*

To accelerate dataset iteration and ensure high-quality corpora for pretraining, we develop an automated evaluation framework that integrates both offline and online assessments. This design enables rapid feedback on new data versions, supporting fast-paced development cycles under limited computational budgets.

**Offline Evaluation**. We conduct evaluations across multiple domains, tailoring strategies to the characteristics of each data type. For natural language corpora—including English, Chinese, and Southeast Asian (SEA) languages—we assess both diversity and quality. Diversity is measured through lexical diversity metrics, which capture vocabulary variation, phrasing richness, and cross-lingual coverage. Quality is assessed using LLM-based judges with customized rubrics, scoring data on fluency, informativeness, and naturalness. For code and mathematics, we focus on syntactic correctness, functional validity, and logical consistency, while also tracking the proportion of low-quality instances to provide deeper insight into noise distribution.

**Online Evaluation**. To measure the downstream impact of data composition, we adopt a lightweight 1.3B parameter proxy model trained on 150B tokens. This experimental setup allows systematic comparison of models pretrained on alternative data mixtures, revealing the relative contributions of multilingual enrichment, domain-specific filtering, and noise reduction. Online results provide early indicators of dataset effectiveness in real-world training scenarios, complementing offline quality signals.

Together, these offline and online evaluations establish a fast, iterative loop that guides corpus construction. By reducing reliance on full-scale pretraining runs, our framework enables efficient exploration of data versions, ensuring that only the most promising mixtures are promoted to large-scale training.

## 3. Instructions Tuning

While general-purpose instruction tuning achieves broad coverage, vertical domains such as e-commerce introduce distinctive challenges. In Southeast Asia (SEA), these difficulties are amplified: models must support heterogeneous workflows (e.g., product search, recommendation, customer service, compliance) while handling a linguistically diverse user base spanning multiple low-resource languages. Existing open-source efforts under-represent both vertical specificity and robust multilingualism.

In this part, we propose an optimized instruction-tuning approach for large-scale Mixture-of-Experts (MoE) models tailored to SEA e-commerce. Our contributions are threefold: (i) **an e-commerce–enhanced training pipeline** that preserves general-purpose ability while substantially improving vertical performance; (ii) **a large-scale e-commerce instruction dataset**, systematically constructed via a closed-loop data flywheel and enriched with synthetic SEA multilingual instructions, designed to capture diverse and business-critical workflows beyond what general-purpose corpora provide; and (iii) **framework-level optimizations** tailored to SFT, enabling efficient adaptation of MoE models.

### 3.1. E-commerce–enhanced Training Pipeline

In Compass-v3, we unify the two-stage SFT pipeline of Compass-v2 (Maria, 2025) into a single stage as shown in Figure 8, which yields consistently stronger performance. We hypothesize that this improvement arises from the model's greater learning capacity: multi-stage fine-tuning risks overfitting to narrow domain-specific data in later stages. Conceptually, however, our data distribution still reflects two components: (i) general instruction-following and (ii) deep reasoning with e-commerce and multilingual specialization.

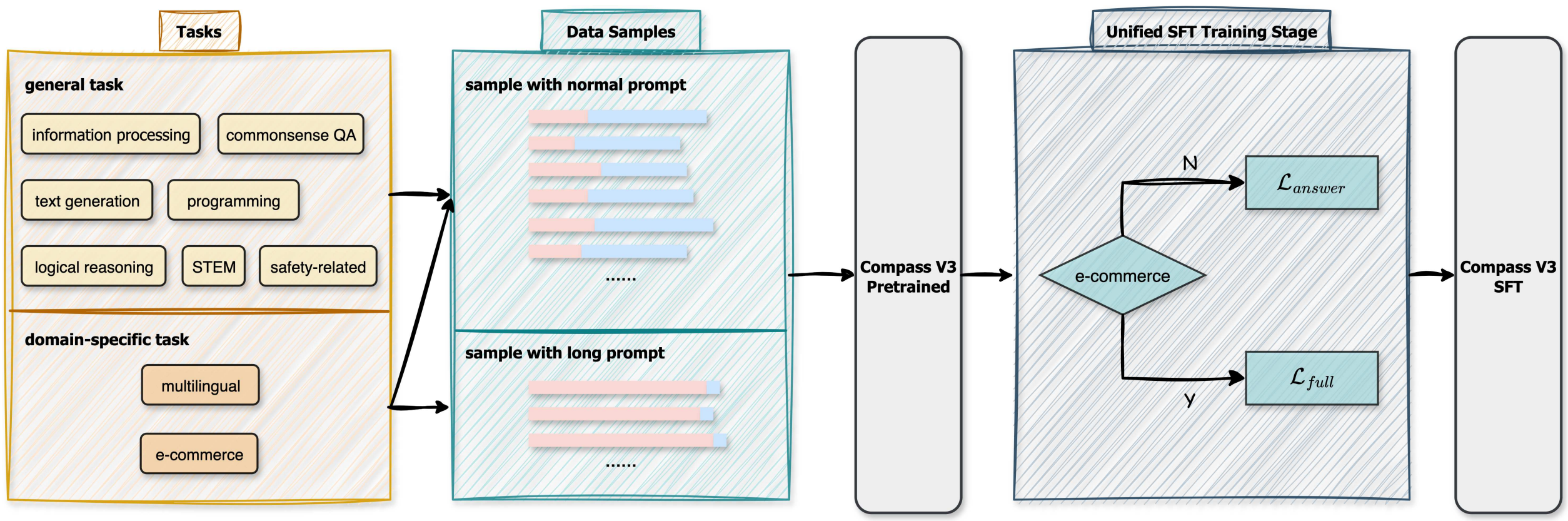

Figure 8 | Overview of the Compass-v3 supervised fine-tuning (SFT) pipeline. General tasks (e.g., commonsense QA, STEM, programming) and domain-specific tasks (multilingual and e-commerce) are converted into data samples with either short or long prompts. The unified SFT stage applies a task-dependent loss: answer-only supervision for general tasks, and full-sequence supervision for e-commerce tasks with long, information-rich prompts. This mixed-loss design strengthens both general instruction following and domain-specific comprehension in e-commerce and multilingual scenarios.

In the **first part**, we aimed to establish strong generalization and instruction-following abilities. The model was fine-tuned on a diverse set of datasets spanning multiple domains, including information processing, commonsense QA, text generation, programming, logical reasoning, STEM, and safety-related tasks. This foundation enabled the model to effectively understand and execute a wide range of instructions.

In the **second part**, we introduce e-commerce and multilingual data to enhance vertical expertise. These tasks often feature extremely long prompts—hundreds or thousands of tokens rich in context—paired with short answers. Conventional answer-only loss discards much of this contextual signal. To mitigate this, we adopt a **mixed-loss strategy**: for e-commerce tasks, we computed loss over the entire sequence, while for other tasks we retained the answer-only loss.

For a training sample consisting of a prompt $x$ and an answer $y$, the standard SFT loss only considers the answer tokens:

$$\mathcal{L}_{\text{answer}} = -\sum_{t=1}^{|y|} \log P_\theta(y_t \mid x, y_{<t}), \quad (1)$$

where $P_\theta$ is the model distribution parameterized by $\theta$.

For e-commerce tasks, we compute the loss over the entire sequence (prompt + answer):

$$\mathcal{L}_{\text{full}} = -\sum_{t=1}^{|x|+|y|} \log P_\theta(s_t \mid s_{<t}), \quad (2)$$

where $s = (x, y)$ denotes the concatenation of prompt and answer.

Finally, we combine the two strategies in a task-dependent manner:

$$\mathcal{L} = \begin{cases} \mathcal{L}_{\text{full}}, & \text{if sample} \in \text{e-commerce domain}, \\ \mathcal{L}_{\text{answer}}, & \text{otherwise}. \end{cases} \tag{3}$$

Empirically, this mixed-loss design bridges general-purpose and vertical training, enabling the model to excel at instruction following while demonstrating stronger comprehension in complex e-commerce and multilingual workflows.

Finally, Compass-v3 extends the **hybrid reasoning** paradigm of Compass-v2, jointly supporting deep reasoning and broad task adaptability. Unlike approaches that separate reasoning-oriented and general-purpose models, our unified design avoids fragmentation, allowing a seamless user experience across reasoning-intensive and everyday tasks without sacrificing flexibility.

## 3.2. Data

### *3.2.1. Large-scale E-commerce Instructions*

E-commerce applications encompass a wide range of complex and heterogeneous tasks, including product categorization and search, personalized recommendation, customer service dialogue, and compliance review, etc. General-purpose instruction-tuning datasets, while effective for broad applications, are insufficient to address these domain-specific requirements. To bridge this gap, we construct a large-scale dataset tailored for real-world e-commerce scenarios. Trained on this dataset, our model Compass-v3, exhibit strong performance across both open source benchmarks and internal e-commerce evaluations.

To ensure robustness and practical utility, the dataset construction follows three core principles:

- **Diverse Instruction Formats:** Covering multiple languages and heterogeneous formats to ensure the model can robustly handle inputs across regions and user styles.
- **Real-world Business Scenarios:** Developed in close collaboration with Shopee business units to guarantee alignment with practical application needs.
- **High-quality Responses:** Strict validation through a combination of human annotation, multi-dimensional filtering, and LLM-based evaluation.

Building upon these principles, the dataset is systematically organized under the Compass-v2 taxonomy, which adopts a three-level hierarchy with 5 second-level categories and more than 70 fine-grained third-level tasks. This taxonomy captures the essential e-commerce workflows (e.g., product search, recommendation, customer service, compliance review), ensuring both breadth and depth in vertical-domain coverage.

To operationalize this construction process at scale, we introduce the **data flywheel** as the central mechanism for dataset generation, as shown in Figure 9. The flywheel enables continuous expansion and refinement by integrating automated selection with human validation, thereby balancing scalability with quality assurance.

- **Data Sources.** Human-curated datasets from the post-training data team, business requirement surveys, and request logs from model-serving platforms.
- **High-value Task Discovery & Long-tail Coverage.** Priority is given to capturing both business-critical tasks (high frequency) and underrepresented long-tail cases (e.g., niche product categories, multilingual complaints). Alignment with the Compass-v2 taxonomy ensures systematic discovery of novel and high-value tasks.

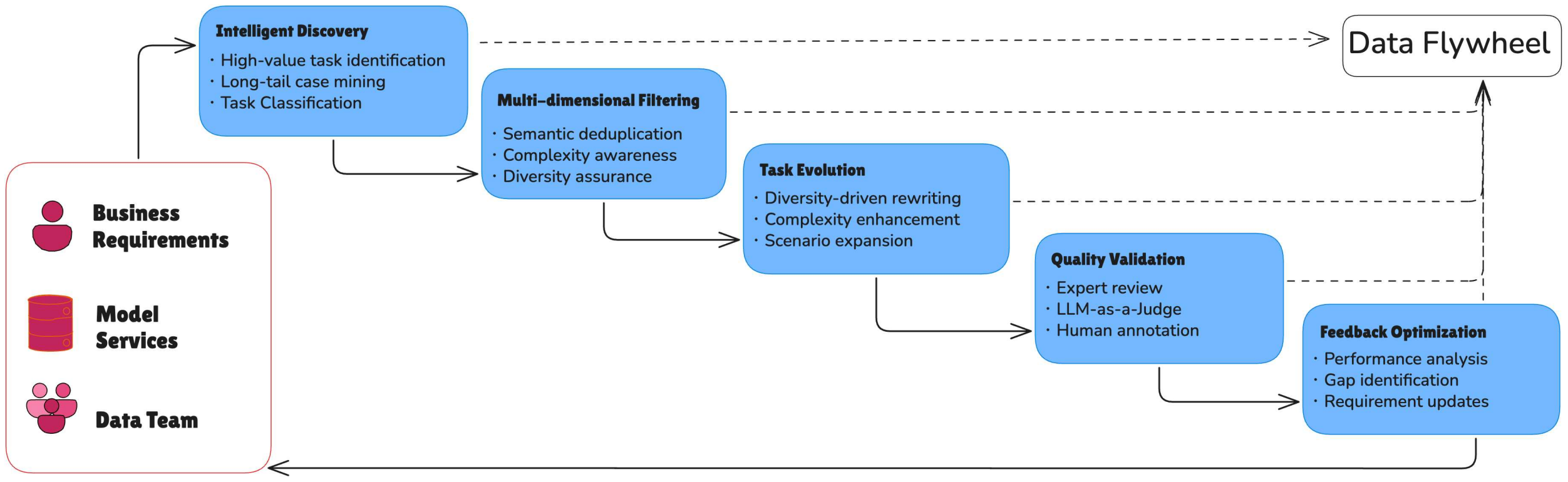

Figure 9 | The data flywheel pipeline for constructing large-scale e-commerce instruction datasets. Business requirements and model feedback drive task discovery, filtering, and evolution, followed by multi-stage quality validation. Deployment feedback is reintegrated to close the loop, enabling continuous expansion and refinement while balancing scalability with quality assurance in real-world e-commerce scenarios.

- **Multi-dimensional Filtering.** Semantic similarity checks, de-duplication of instructions and responses, and complexity-aware filtering are applied to preserve diversity while retaining tasks requiring multi-step reasoning and contextual understanding.
- **Task Rewriting & Complexity Evolution.** Leveraging diversity-driven paraphrasing to enrich lexical and syntactic variety, while complexity evolution transforms simple tasks into more challenging scenarios (e.g., single-turn → multi-turn dialogue, single-entity → multi-entity reasoning, single-instruction → multi-instruction). This enhances generalization and reasoning capabilities.
- **Feedback & Continuous Improvement.** Real-world deployment feedback, user interaction logs, and evaluation results are systematically analyzed to identify failure cases and performance gaps. These insights feed back into data sourcing, filtering, and task design, thereby closing the loop and enabling a self-reinforcing flywheel.

In summary, the proposed data flywheel pipeline establishes a self-reinforcing closed loop. Beginning with data sourcing and task discovery, proceeding through multi-dimensional filtering and complexity evolution, and culminating in validation and feedback integration, the pipeline enables both quality assurance and continuous refinement. Manual annotation, prompt engineering, and multi-stage evaluation (including domain-expert review and LLM-as-a-Judge assessment) are embedded as critical checkpoints to ensure reliability. The iterative feedback mechanism further drives the expansion and adaptation of the dataset, thereby supporting sustained improvement of downstream e-commerce applications.

#### 3.2.2. *SEA Multilingual Instructions*

While existing open-source LLM research has predominantly focused on English and Chinese language processing, comprehensive multilingual support for real-world applications remains largely unexplored. As a leading e-commerce platform serving diverse Southeast Asian markets, Shopee requires robust multilingual capabilities that extend beyond traditional language pairs to encompass the linguistic diversity of its regional user base.

To address this gap, we construct a comprehensive instruction-tuning dataset of approximately **2.81 million** instances, specifically designed for multilingual applications. The dataset primarily targets six additional languages beyond Chinese and English: Malay (ms), Thai (th), Portuguese (pt), Vietnamese (vi), Tagalog (tl), and Indonesian (id), to better align with Shopee's markets. The dataset is constructed from four sources:

- **Data from open-source instruction datasets.** Our multilingual dataset incorporates data from open-source instruction corpora, spanning high-resource languages such as English and Chinese, as well as low-resource languages including Southeast Asian languages and others. We filter these datasets to focus on our target languages to improve the model's performance on them.
- **Data constructed from open-source non-instruction datasets.** Due to the limited availability of multilingual instruction datasets, we also construct instruction-tuning data from non-instruction datasets such as translation and summarization corpora. We augment these datasets with various prompts to make them suitable for instruction fine-tuning.
- **Data translated from open-source instruction datasets.** To further expand the dataset, we translate a portion of instruction-related data, ensuring the model's ability to follow instructions across multiple languages.
- **Multi-agent generated data.** In several open-source datasets, we observed that while prompts are often well-designed, the quality of responses remains suboptimal. To address this, we adopt a Multi-Agent Optimization (MOA) framework, where multiple agents collaborate to generate and refine responses. This process distills higher-quality outputs, enhancing their reliability and overall performance.

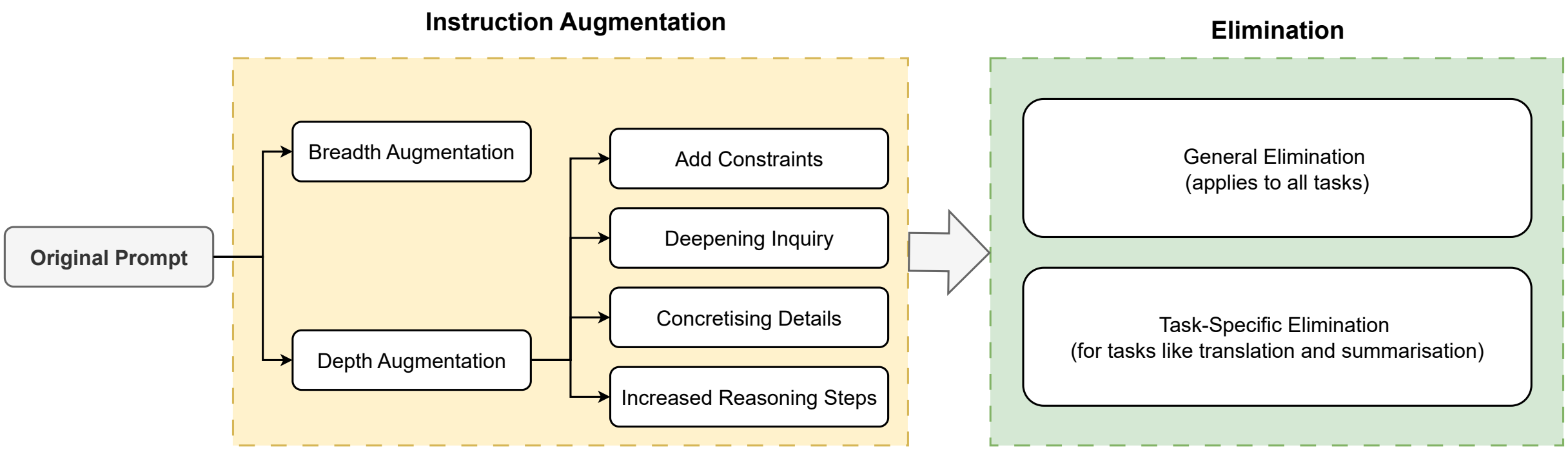

Figure 10 | Multilingual instruction augmentation pipeline. Original prompts are expanded along two dimensions: breadth, through domain-consistent alternatives, and depth, via added constraints, concretized details, reasoning requirements, or broader inquiry. Generated variants then pass through elimination stages—general filtering for coherence and value, and task-specific rules (e.g., translation, summarization)—to ensure fidelity and quality.

This collection provides a broad foundation but remains insufficient in capturing the diversity of instruction formats and the business-grounded multilingual task scenarios required in practice.

To overcome these limitations, we build the instruction-evolution framework based on WizardLM Xu et al. (2024) and design an instruction augmentation pipeline (as shown in Fig. 10). The pipeline evolves prompts into more challenging variants along two dimensions: *breadth*, by generating domain-consistent alternatives, and *depth*, by adding constraints, concretizing vague instructions, requiring reasoning, or broadening inquiry. Evolved instructions then pass through a general elimination stage to ensure added value, coherence, and language consistency. For task-specific cases such as summarization and translation, we introduce additional filtering rules to preserve input fidelity and prevent invalid outputs, such as translating into the same language as the source.

### 3.3. Framework Optimization

Megatron-LM was originally designed for pre-training, and although we adapted it for SFT in Compass-V2, several limitations remained. To address these issues, we made the following improvements:

- **Complete data packing** Instead of truncating the final sample during packing, we append pad tokens at the end. This prevents incomplete samples, which are common in pre-training but problematic in SFT due to the presence of special tokens. Incomplete samples can distort the learning distribution and lead to meaningless truncated outputs. Our revised packing strategy ensures that if the maximum sequence length cannot accommodate a full sample, padding is used rather than forcing a partial one.
- **Refined attention masking** In Megatron-LM, conversation turns are typically separated by an eos token to generate attention masks. However, this design prevents later turns from attending to earlier context in multi-turn dialogues. To resolve this, we introduced an eot token between dialogue turns, ensuring that the attention mask is generated correctly and that inter-turn dependencies are preserved.

## 4. Reinforcement Learning

Reinforcement Learning for Southeast Asian e-commerce presents unique challenges: multilingual variability, rich product semantics, and nuanced user expectations that conventional pipelines—largely developed for high-resource, general-purpose tasks—struggle to capture. We propose an efficient framework that departs from costly large-scale online RL and instead combines **(i) Mixed-Policy Reinforcement Learning**, which flexibly integrates on- and off-policy signals across domains, (ii) **Optimal Transport Preference Optimization (OTPO)** (Li et al., 2025) for token-level alignment, explicitly emphasizing preference-critical tokens while reducing noise from irrelevant content, and (iii) **domain-specialized reward models with hybrid RM+LLM verification**. To further improve efficiency, we precompute reference log-probabilities, cutting training cost by 16.6% without sacrificing stability. Our approach achieves state-of-the-art alignment on multilingual and e-commerce benchmarks, offering a scalable and resource-conscious path toward real-world deployment in low-resource Southeast Asian markets. In future work, we will extend this framework with large-scale online RL strategies to further enhance adaptability and robustness.

### 4.1. Mixed-Policy Reinforcement Learning

A central challenge in aligning large language models for e-commerce and multilingual tasks is balancing the efficiency of off-policy data with the fidelity of on-policy exploration. To this end, we adopt a **mixed-policy Reinforcement Learning** strategy, which integrates both paradigms in a domain-adaptive manner, as shown in Fig. 11.

The key intuition is that the *size of the output space* determines the relative importance of on- versus off-policy rollouts. For domains with a small exploration space (e.g., structured agent calls or simple instruction following), high-quality off-policy responses suffice, as deviations from the model's distribution are limited. Conversely, for domains with large exploration spaces (e.g., e-commerce reasoning or multilingual generation), on-policy rollouts are essential for surfacing failure modes and providing training signals that reflect the model's actual weaknesses.

Formally, for each prompt $x$, we generate:

- **Rejected responses** exclusively from the on-policy model $\pi_\theta$, ensuring that negative samples reflect the current policy's behavior.

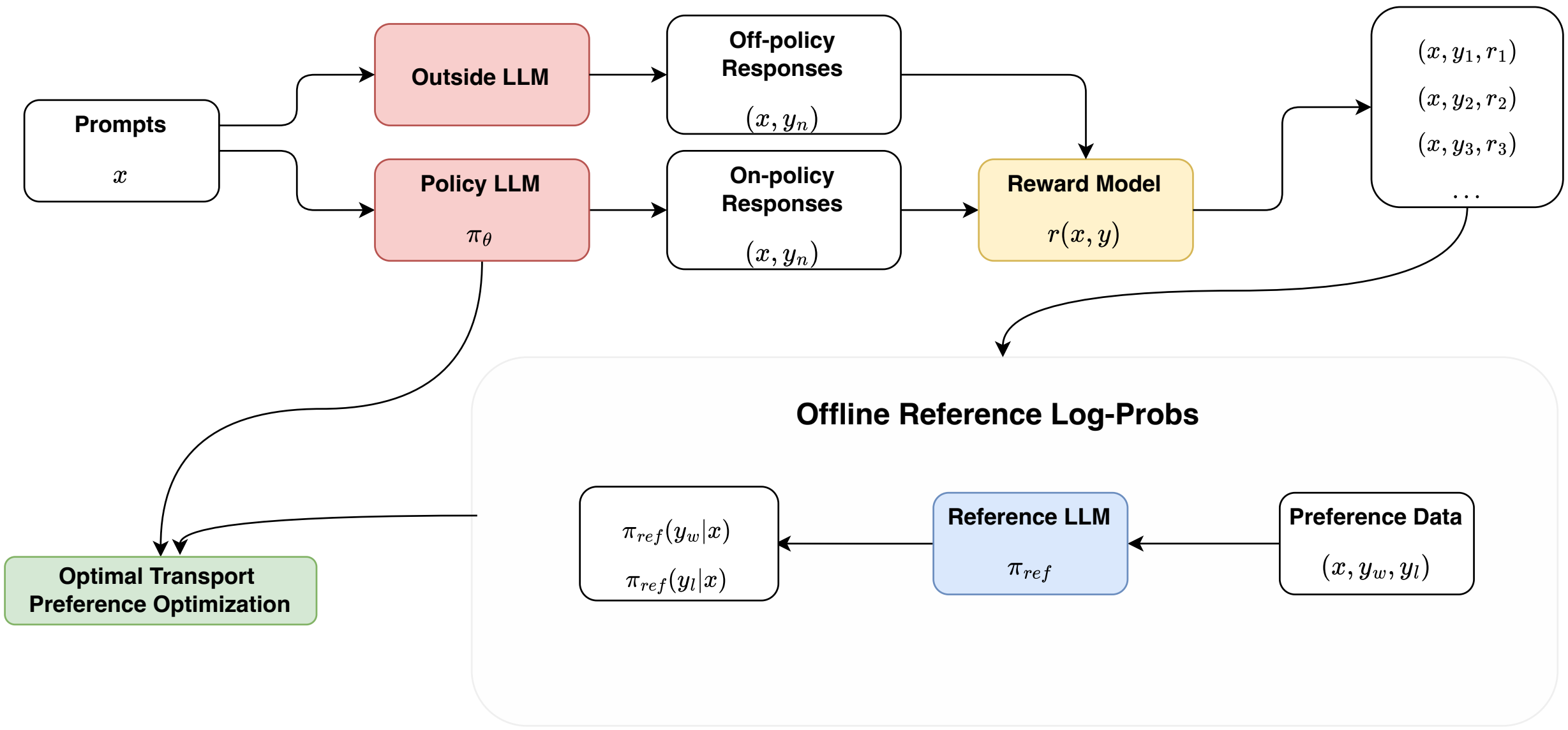

Figure 11 | Overview of our alignment framework. A policy LLM generates on-policy responses, while advanced LLMs contributes off-policy samples of higher quality. Preference data are evaluated through a domain-specialized reward model, and *offline reference log-probabilities* are precomputed to reduce inference overhead during training. Final optimization is performed using **Optimal Transport Preference Optimization (OTPO)**, which emphasizes preference-critical tokens for fine-grained alignment. This design jointly improves efficiency and effectiveness for reinforcement learning in multilingual e-commerce settings.

- **Chosen responses** from a mixture of sources, combining off-policy responses from a stronger reference model $\pi_{\text{ref}}$ with on-policy samples. This hybrid construction yields positive examples that are both high-quality and distributionally relevant.

This mixed-policy design improves coverage of realistic error modes while maintaining data efficiency, thereby stabilizing downstream optimization. Empirically, it produces preference pairs that are more diverse and semantically rich, enabling subsequent token-level optimization to focus on preference-critical distinctions.

Table 1 | Output space size and policy suitability across domains. Smaller, structured tasks (agents, instruction following) benefit from off-policy data, whereas larger, more diverse tasks (e-commerce, multilingual generation) require on-policy exploration. This motivates our mixed-policy RL strategy.

| Domain | Output Space Size | Characteristics | Suitable Policy |
|---|---|---|---|
| Agent | Small | Structured APIs | Off |
| Instruction Following | Small–Medium | Task-focused | Off |
| E-commerce | Medium–Large | Rich product semantics | On |
| Multilingual | Large | High variability in languages | On |

### 4.2. Optimal Transport Preference Optimization (OTPO)

In Southeast Asian e-commerce and multilingual applications, conventional reinforcement learning algorithms often fail to capture nuanced user expectations due to large and diverse output spaces.

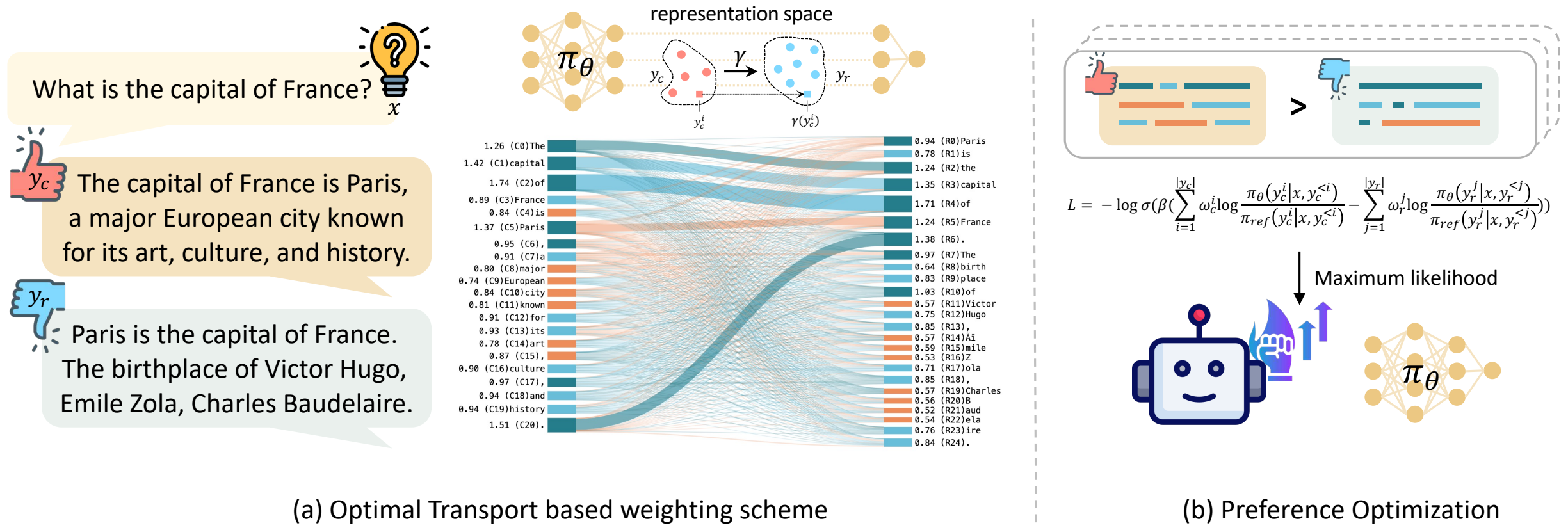

Figure 12 | Overall framework of OTPO. (a) We compute the token-level weighting scheme using optimal transport. Each response's distribution is made up of its tokens, represented as vectors in the LLM's representation space. The optimized transport plan is visualized using a Sankey diagram. (b) We decompose the DPO loss at the token level and apply the weighting scheme obtained in (a).

This motivates us to go beyond coarse response-level alignment and design a token-sensitive method that can emphasize preference-critical distinctions while suppressing noise. To this end, we introduce **Optimal Transport Preference Optimization (OTPO)** (Li et al., 2025) as shown in Fig. 12, which integrates seamlessly with our broader framework of mixed-policy reinforcement learning and domain-specialized reward modeling. OTPO explicitly leverages the geometric structure between chosen and rejected responses, making it particularly suited for complex multilingual and commerce-specific generation tasks.

Table 2 | Overall pairwise accuracy on Shopee-internal multilingual and e-commerce evaluations.

| Model | Type | Shopee Multilingual Acc | Shopee E-commerce Acc |
|---|---|---|---|
| `gaotang/RM-R1-DeepSeek-Distilled-Qwen-7B` | GRM | 0.5354 | 0.6000 |
| `gaotang/RM-R1-DeepSeek-Distilled-Qwen-14B` | GRM | 0.6768 | 0.6118 |
| `skyworkv2-qwen3-8b` | Scalar | 0.8182 | 0.7265 |
| `Qwen-3-Nemotron-32B-Reward` | Scalar | 0.7677 | 0.6118 |
| `multilingual-sorted` | Scalar | 0.8182 | 0.7471 |
| `multilingual-unsorted` | Scalar | 0.8182 | 0.7500 |
| `multilingual-ecommerce-sorted` | Scalar | **0.8485** | **0.7559** |

### 4.3. Domain-specific Reward Model

Designing effective reward models for reinforcement learning in Southeast Asian multilingual and e-commerce applications requires domain-specific strategies beyond generic pipelines. We introduce a hybrid framework that combines **rule-based LLM verification with a scalar reward model (RM)** tailored to commerce and multilingual data. The RM is initialized from `Skywork-v2` (Liu et al., 2025a) but enhanced via curated Shopee preference pairs and curriculum ordering based on semantic deviation. Training uses a margin-augmented Bradley–Terry (Bradley and Terry, 1952; Touvron et al., 2023) objective with head re-initialization, which strengthens preference separation and reduces tie rates. Experiments(Tab. 2) on Shopee-internal benchmarks show that our approach outperforms generative RMs (Guo et al., 2025) and vanilla scalar baselines (Liu et al., 2025a; Wang et al., 2025), with +3.0% accuracy in multilingual and +2.9% in e-commerce settings. Ablation studies confirm the effectiveness of margin tuning and curriculum sorting. ***These results demonstrate that domain-specific RM design—combining structured rule-based signals and preference-optimized scalar***

***modeling—provides a robust foundation for alignment in real-world multilingual and commercial applications***.

# 5. Inference Optimization

## 5.1. Bottlenecks

MoE architectures are widely adopted to increase model capacity without proportionally raising computational cost, as only a small subset of experts is activated per token (Fedus et al., 2022). However, in our model, we adopt a "fewer-but-larger" expert design: although each input activates only 4 experts out of 16, each expert is itself a large Transformer block with substantial hidden dimensions. This differs fundamentally from smaller experts used in DeepSeek V3, where communication overhead dominates.

Our profiling study reveals that the inference bottleneck in Compass is instead computational. Figure 13 shows the breakdown of forward latency, where grouped GEMM operations in expert feed-forward layers account for over 55% of runtime. Router overhead and expert communication contribute far less (below 15%). This implies that communication optimizations cannot meaningfully improve decoding speed. Instead, reducing the *compute intensity* of expert GEMMs is the key to achieving real-time inference throughput.

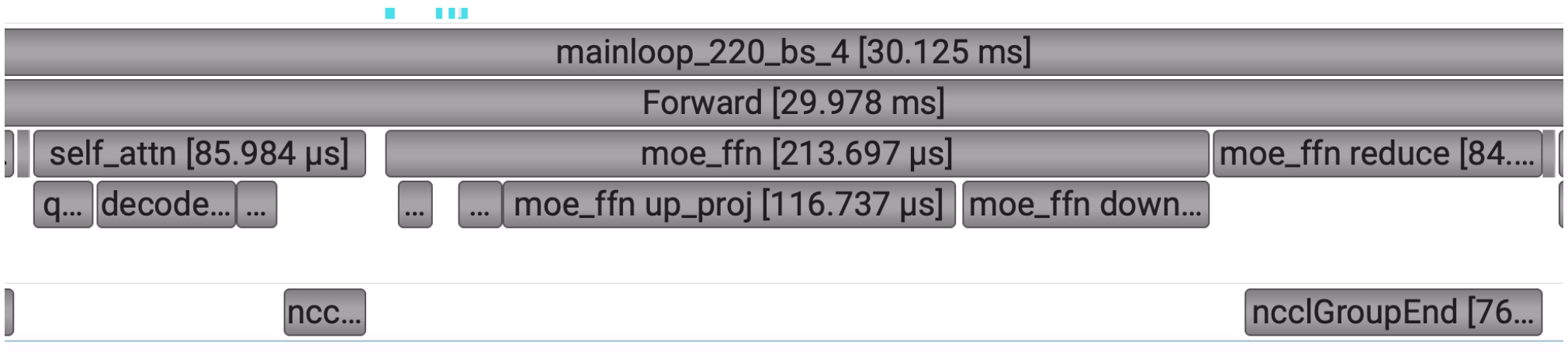

Figure 13 | Profiling of Compass v3 forward pass (batch size = 4). MoE feed-forward GEMMs dominate runtime, while self-attention and communication incur minimal cost, confirming that inference is compute-bound and motivating expert-aware quantization.

## 5.2. Expert-Aware Quantization

To alleviate this computational bottleneck, we introduce an **expert-aware FP8(W8A8) quantization framework**, as illustrated in Figure 14 tailored for large MoE inference. While FP8 quantization has recently been demonstrated to accelerate dense LLMs by leveraging Hopper tensor cores, directly applying it to MoE is ineffective. MoE models exhibit highly heterogeneous activation distributions: some experts are frequently activated and receive diverse token contexts, while others fire rarely and are under-represented in calibration. As a consequence, the direct computation of FP8 activation scaling factors results in underfitting of rare experts, leading to substantial accuracy degradation.

Our method addresses this through two complementary strategies:

1. **Balanced Expert Calibration.** We rebalance the calibration dataset by oversampling additional tokens to rarely activated experts through router configuration analysis. This ensures each expert has sufficient statistical coverage for reliable scaling estimation.
2. **Unified Channel-wise Smoothing.** Activation smoothing (Xiao et al., 2023) is effective for suppressing outliers, but applying it naively per-expert introduces dynamic overhead. Instead, we construct a single channel-wise smoothing vector derived from both experts and router weight, capturing the maximum scaling requirements while avoiding runtime re-scaling.

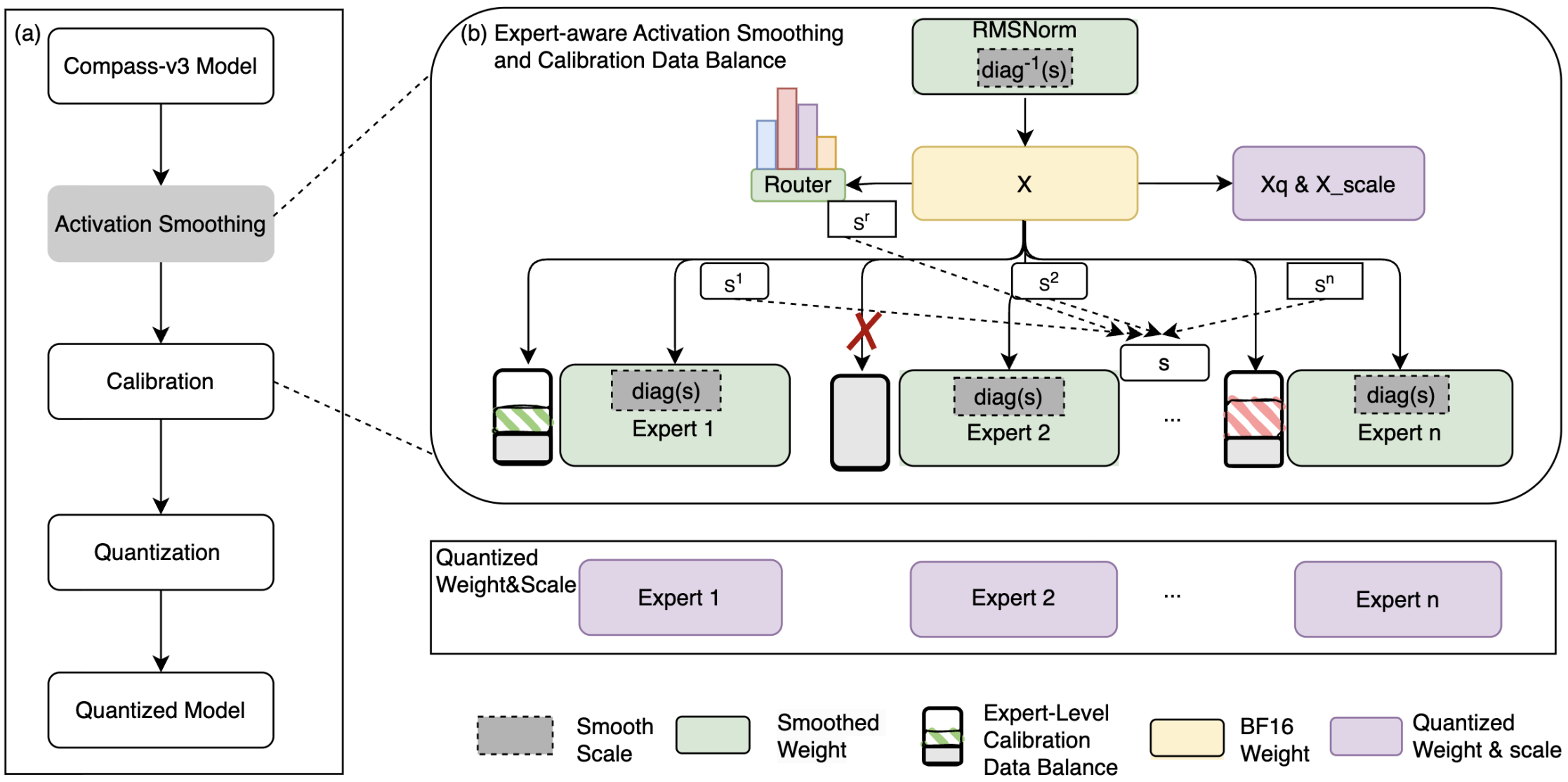

Figure 14 | (a) Expert-aware FP8 quantization framework. Compared with naive FP8 quantization, we additionally perform expert-aware activation smoothing and adjust the calibration data distribution at the expert level. (b) Expert-aware activation smoothing methods suppresses activation outliers by applying a unified smoothing vector s across both MoE experts and the router. For activations X, we fuse s into the preceding RMSNorm layer through parameter fusion. To address the inherent imbalance of expert activations in Compass-v3, we oversample calibration data for under-represented experts until their activation counts satisfy a predefined criterion, thereby improving the accuracy of quantization parameter estimation.

These techniques jointly mitigate the mismatch between FP8 quantization and MoE sparsity, enabling accurate and efficient inference. Unlike weight-only quantization approaches such as AWQ (Lin et al., 2024), which introduce extra dequantization overhead, our framework fully exploits Hopper FP8 tensor cores for end-to-end GEMM acceleration.

### 5.3. Results and Throughput Gains

We benchmark our approach on multilingual and e-commerce evaluation suites derived from Compass datasets. Expert-aware FP8 consistently delivers **2× speedup** over FP16 baselines, sustaining up to **1,000 tokens/s** on 8*H100 cards. Importantly, as batch size increases, the performance gap widens, indicating strong scalability under production workloads.

Fig. 15 reports accuracy changes relative to the BF16 baseline across seven multilingual benchmarks. We observe that vanilla FP8 quantization introduces substantial degradation, particularly in low-resource languages such as Thai (−10.17%), Vietnamese (−11.75%), and Chinese (−7.37%). In contrast, our expert-aware FP8 scheme consistently mitigates these losses, reducing degradation to within −2.5% across all languages. Notably, performance even surpasses BF16 in Portuguese (+1.05%), indicating that expert-aware calibration not only preserves model fidelity but can also enhance robustness. These results confirm that computation-centric, expert-aware quantization is critical for scaling MoE inference to multilingual applications without sacrificing quality.

These results demonstrate that **expert-aware FP8 quantization is essential for MoE inference**.

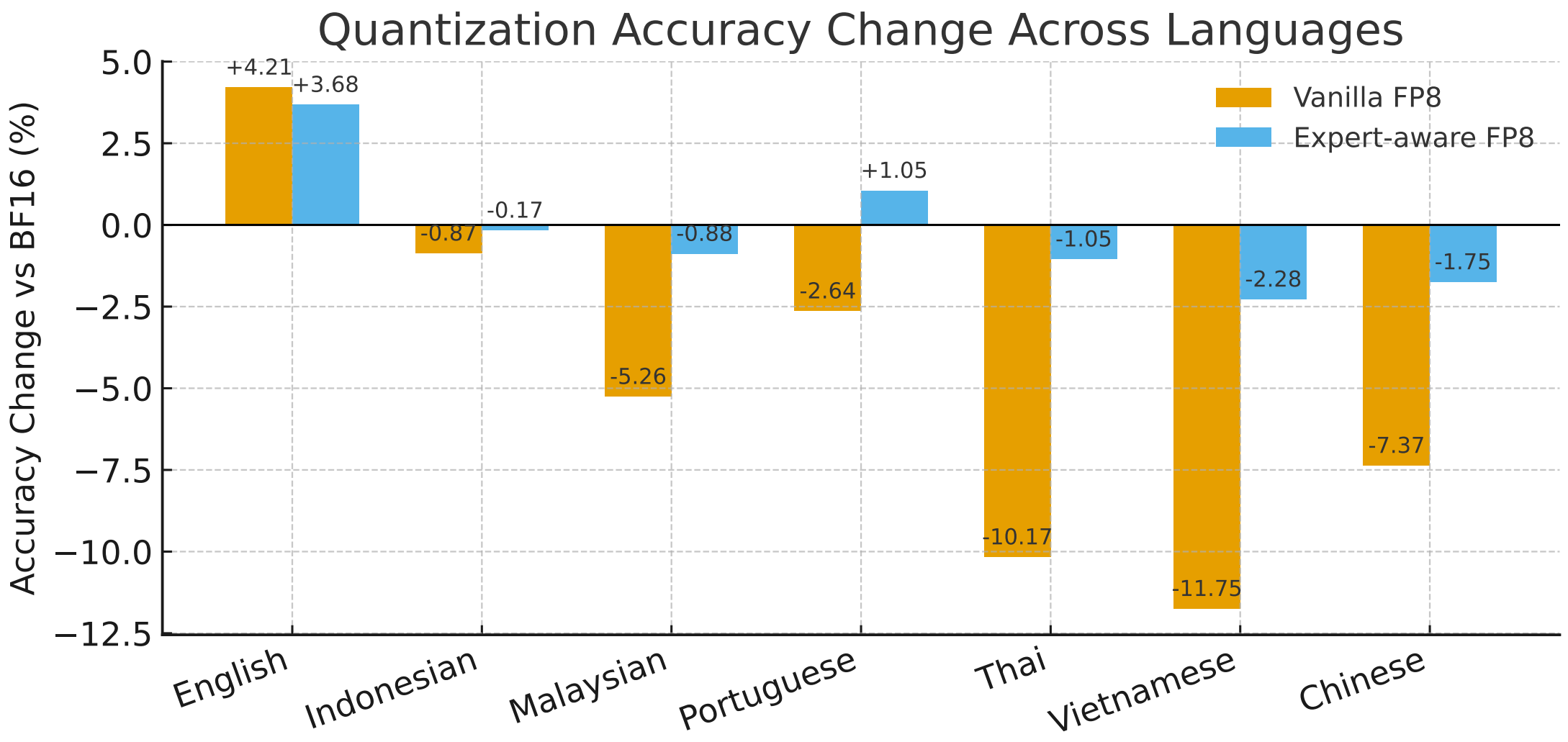

Figure 15 | Accuracy change on in-house multilingual benchmarks relative to the BF16 baseline. Vanilla FP8 introduces notable degradation across several languages, especially Thai, Vietnamese, and Chinese, whereas expert-aware FP8 substantially reduces the loss and achieves near-parity performance.

By adapting quantization techniques to expert sparsity and activation heterogeneity, we show that large experts can be deployed efficiently at scale without compromising model fidelity.

## 6. Evaluation

Large Language Models (LLMs) are commonly evaluated with general-purpose benchmarks, yet these fail to capture domain-specific strengths in Southeast Asian (SEA) e-commerce. We present a systematic evaluation of Compass-v3, a large-scale Mixture-of-Experts model built for Shopee's multi-market ecosystem. Our framework combines open-source datasets with a newly constructed **in-house e-commerce benchmark**. For standard capabilities, we adopt the llm-evaluation-harness framework (Gao et al., 2024) with automated judgments and selective manual review. For domain-specific, open-ended tasks, we employ GPT-4.1 as an evaluator to ensure fairness and consistency.

### 6.1. Evaluation Setting

#### *6.1.1. Basic Setting*

To demonstrate the evaluation process, we list the hyperparameters used for each models.

- **Compass-v3**: max_tokens=2048, seed=42, repetition_penalty=1.05, top_k=5, top_p=0.85, temperature=0.4
- **Qwen Series**: repetition_penalty=1.05, temperature=0.7, top_p=0.8, top_k=20
- **GPT Series/DeepSeek V3.1**: official API defaults

#### *6.1.2. In-House Dataset Construction*

Shopee is the leading e-commerce platform in Southeast Asia, covering more than ten markets including Singapore, Malaysia, the Philippines, Thailand, Vietnam, and Brazil. Domain-specific capabilities in e-commerce are what Compass-v3 values most. However, existing evaluations of LLMs mainly focus on general capabilities, which are insufficient to reflect performance in the e-commerce domain. There are some open-source e-commerce evaluation datasets, such as Shopping-MMLU, eCellm, and ChineseEcomQA. The main tasks in these datasets include product title generation, product attribute extraction, similar product identification, and product recommendation based on user behavior sequences. However, datasets derived from real e-commerce business scenarios—such as product guidance, after-sales service, and product understanding—are extremely limited. Moreover, existing open-source e-commerce datasets are only in English and Chinese, with no inclusion of Southeast Asian languages. To address these issues, we construct an in-house e-commerce evaluation dataset, primarily sourced from real business scenarios. This dataset covers seven languages: English, Chinese (both Traditional and Simplified), Indonesian, Vietnamese, Thai, Portuguese, and Malay.

### 6.2. In-House Ecommerce Benchmark Evaluation

**Table 3 reports model performance on our in-house e-commerce benchmark across five major domains. Compass-v3 consistently achieves the highest overall accuracy (86.56%)**, surpassing both proprietary frontier models (e.g., GPT-4.1 at 82.03%, GPT-4o at 76.67%) and strong open-source baselines (e.g., DeepSeek-V3.1 at 76.55%, Qwen3-235B-A22B at 79.00%). Notably, Compass-v3 shows clear advantages in practical business scenarios such as Shopping Guide (96.83%), After-Sales Issue resolution (96.45%), and Product Type Understanding (97.32%), reflecting its alignment with real-world e-commerce needs. While GPT-4.1 exhibits competitive performance in generative tasks such as Product Info Generation (90.70%), Compass-v3 maintains balanced strengths across both discriminative (e.g., Query Understanding, Product Classification) and generative (e.g., Title Generation, Modification) tasks. These results demonstrate that Compass-v3 not only rivals general-purpose LLMs but also delivers domain-specialized capabilities essential for SEA e-commerce applications.

Table 3 | In-House Ecom Dataset Evaluation

| domain | ecom tasks | **Compass-v3** | **DeepSeek-V3.1** | GPT-4.1 | GPT-4.1-mini | GPT-4o | GPT-4o-mini | GPT-oss-120B | Qwen3-235B-A22B |
|---|---|---|---|---|---|---|---|---|---|
| | Shopping Guide | 96.83 | 86.51 | 87.78 | 83.13 | 88.25 | 77.74 | 81.43 | 88.61 |
| | Brand Knowledge | 88.36 | 78.84 | 76.19 | 84.49 | 78.31 | 71.95 | 75.13 | 78.88 |
| | After Sales issue | 96.45 | 72.34 | 90.31 | 72.54 | 78.01 | 69.21 | 88.65 | 75.71 |
| | Product Attribute Extraction | 89.93 | 78.16 | 82.84 | 76.8 | 78.44 | 71.23 | 77.02 | 79.64 |
| | Product Classification | 89.33 | 83.56 | 90.22 | 88.89 | 86.22 | 83.6 | 91.11 | 88.10 |
| | Product Type Understanding | 97.32 | 90.61 | 94.64 | 86.67 | 91.57 | 85.56 | 92.34 | 89.05 |
| | Query Understanding | 91.28 | 77.4 | 86.58 | 77.46 | 79.42 | 75.71 | 81.43 | 81.27 |
| | Query-Product Matching | 73.96 | 53.06 | 58.38 | 51.55 | 52.27 | 58.90 | 47.53 | 53.73 |
| | User Review Understanding | 87.00 | 78.48 | 81.61 | 82.67 | 77.28 | 81.05 | 75.19 | 84.50 |
| | Search Relevance | 82.58 | 76.59 | 77.34 | 79.83 | 73.78 | 73.95 | 80.34 | 80.95 |
| | product recommendation | 76.82 | 66.67 | 70.91 | 64.84 | 63.18 | 65.25 | 64.39 | 67.58 |
| | Similar Product Identification | 96.55 | 86.93 | 92.96 | 81.56 | 87.21 | 80.85 | 85.92 | 85.92 |
| | Product Description Similarity | 81.43 | 88.19 | 81.43 | 86.27 | 85.23 | 85.71 | 87.34 | 88.24 |
| | Product Info Generation | 79.07 | 74.68 | 90.70 | 83.49 | 73.39 | 74.44 | 71.58 | 83.33 |
| | Product Title Generation | 87.18 | 65.06 | 77.56 | 78.96 | 66.67 | 66.34 | 71.79 | 69.58 |
| | Product Title Modification | 70.86 | 67.77 | 73.07 | 76.05 | 67.55 | 72.49 | 71.74 | 68.93 |
| | average | 86.56 | 76.55 | 82.03 | 78.45 | 76.67 | 74.62 | 77.68 | 79.00 |

### 6.3. Open-source Ecommerce Benchmark Evaluation

Considering the fairness of evaluation, we carefully evaluated the model's capabilities on open-source e-commerce datasets (i.e., Shopping MMLU and ECInstruct) to verify the practical applicability of the model. From Table 4, we can conclude that **Compass-v3 demonstrates comprehensive strengths in e-commerce related tasks, ranking among the top three models on both benchmarks.**

Specifically, Compass-v3 achieves 49.50% on Shopping MMLU, outperforming all baselines, including GPT-4.1 (45.63%), Qwen3-235B-A22B (44.76%), and GPT-oss-120B (41.93%). This suggests that Compass-v3 is particularly effective in handling product- and shopping-related knowledge, even when benchmarked against larger proprietary models. On the ECInstruct test set, Compass-v3 attains a strong score of 85.14%, comparable to top-performing systems such as GPT-4.1 (86.70%) and GPT-4o (85.61%), better than other models such as GPT-4.1-mini (78.28%) and GPT-4o-mini (79.81%). These results highlight Compass-v3's ability to perform well in e-commerce tasks, validating its practical utility in online shopping scenarios.

Table 4 | E-commerce Evaluation

| Benchmark | Compass-v3 | DeepSeek-V3.1 | GPT-4.1 | GPT-4.1-mini | GPT-4o | GPT-4o-mini | GPT-oss-120B | Qwen3-235B-A22B |
|---|---|---|---|---|---|---|---|---|
| ECInstruct (test set) | 49.50 | 41.59 | 45.63 | 38.45 | 42.78 | 37.94 | 41.93 | 44.76 |
| Shopping MMLU | 85.14 | 82.98 | 86.70 | 78.28 | 85.61 | 79.81 | 83.96 | 83.20 |
| **Average** | 67.32 | 62.29 | 66.17 | 58.37 | 64.20 | 58.88 | 62.95 | 63.98 |

## 6.4. In-house Multilingual Benchmark Evaluation

The currently available open-source evaluation datasets mainly consist of multiple-choice and true/-false questions, the free-form, open-ended queries commonly posed by real users in practical scenarios. To address this, we have constructed a series of in-house evaluation datasets to comprehensively assess the performance of LLMs in multilingual capabilities. Each evaluation question is provided in multiple language versions, including English, Indonesian, Malay, Portuguese, Thai, Vietnamese, and Traditional Chinese. In Table 5 these languages are short for en, id, my, pt, th, vi, tw, respectively. SEA Average refers to the average score across five languages: Malay (my), Portuguese (pt), Thai (th), Vietnamese (vi), and Indonesian (id). We conduct a comprehensive evaluation of the model from a SEA multilingual perspective.

Table 5 | In-House Multilingual Datasets Evaluation

| Language | Compass-v3 | DeepSeek-V3.1 | GPT-4.1 | GPT-4.1-mini | GPT-4o | GPT-4o-mini | GPT-oss-120b | Qwen3-235B-A22B |
|---|---|---|---|---|---|---|---|---|
| en | 87.68 | 88.71 | 89.75 | 88.45 | 87.55 | 87.18 | 85.82 | 88.73 |
| tw | 83.27 | 79.01 | 87.94 | 86.75 | 86.13 | 86.16 | 81.83 | 88.44 |
| id | 85.64 | 84.19 | 87.54 | 86.31 | 85.60 | 86.08 | 83.75 | 86.76 |
| my | 88.51 | 86.93 | 88.42 | 86.86 | 86.92 | 84.93 | 86.00 | 87.31 |
| pt | 89.25 | 80.23 | 88.46 | 86.78 | 86.59 | 87.02 | 87.76 | 87.46 |
| th | 85.77 | 82.76 | 87.64 | 85.63 | 84.70 | 83.04 | 83.60 | 84.83 |
| vi | 84.80 | 82.84 | 85.79 | 83.50 | 81.60 | 81.98 | 82.96 | 83.26 |
| **SEA Average** | 86.79 | 82.66 | 87.57 | 85.82 | 85.08 | 84.61 | 84.81 | 85.01 |

From the evaluation results shown in Table 5, **Compass-v3 achieves a good performance with an SEA language average score of 86.79 across all evaluated languages.** The model shows remarkable consistency, excelling in Indonesian (85.64) and maintaining competitive scores in English (87.68), and Chinese (83.27). Its robust performance in SEA languages like Thai (85.77) and Vietnamese (84.80) further reflects its ability to effectively handle linguistically diverse and low-resource inputs. Notably, Compass-v3 beats all the top models in Portuguese and Malay with a score of 89.25 and 88.51, respectively. Compass-v3's good performance makes it well-suited for real-world applications requiring stable performance across multiple language environments, especially in SEA scenarios.

## 6.5. General Benchmark Evaluation

To comprehensively evaluate Compass-v3, we employ a set of well-established open-source benchmarks that assess diverse capabilities in English, including mathematical reasoning, knowledge-based question answering, reading comprehension, and common sense reasoning. To extend our evaluation to the multilingual setting, we select four representative benchmarks—OpenBookQA, XCOPA, MMLU, and HellaSwag—and translate randomly sampled subsets from each into six languages: Malay, Indonesian, Thai, Vietnamese, Portuguese, and Traditional Chinese. As these multilingual benchmark subsets mainly consist of SEA languages, we denote them with the label "sea" in Table 6.

Table 6 | General Ability Evaluation

| Domain | Benchmark | Compass-v3 | DeepSeek-V3.1 | GPT-4.1 | GPT-4.1-mini | GPT-4o | GPT-4o-mini | GPT-oss-120B | Qwen3-235B-A22B |
|---|---|---|---|---|---|---|---|---|---|
| | GSM8K | 91.28 | 89.27 | 92.15 | 90.23 | 90.69 | 87.41 | 91.24 | 93.00 |
| *Knowledge-based QA* | MMLU | 81.37 | 81.87 | 86.10 | 84.44 | 85.51 | 82.52 | 85.46 | 82.67 |
| | GPQA | 50.00 | 50.74 | 55.90 | 52.22 | 53.60 | 49.43 | 52.14 | 50.50 |
| | ARC | 61.86 | 63.73 | 72.43 | 70.16 | 70.24 | 66.55 | 70.12 | 68.40 |
| *Reading Comprehension & Understanding* | BoolQ | 89.79 | 89.00 | 93.24 | 89.00 | 92.41 | 89.33 | 93.00 | 94.24 |
| | OpenBookQA | 49.20 | 50.91 | 54.35 | 51.47 | 52.23 | 50.91 | 51.77 | 54.64 |
| | RACE | 43.73 | 45.93 | 53.81 | 52.08 | 52.33 | 48.25 | 52.68 | 54.07 |
| | Belebele | 90.89 | 88.40 | 93.24 | 90.61 | 93.28 | 90.07 | 93.70 | 93.70 |
| *Common Sense Reasoning* | HellaSwag | 86.92 | 87.17 | 92.17 | 91.19 | 89.95 | 85.27 | 90.85 | 91.73 |
| | Social IQa | 51.64 | 50.49 | 57.07 | 55.83 | 54.38 | 52.71 | 55.50 | 56.66 |
| | XCOPA | 99.80 | 98.10 | 98.24 | 97.30 | 98.90 | 96.17 | 98.10 | 96.78 |
| | XStoryCloze | 81.47 | 80.06 | 86.12 | 83.37 | 83.45 | 81.97 | 84.17 | 85.10 |
| | Winogender | 87.08 | 85.78 | 93.07 | 89.51 | 89.90 | 83.55 | 90.13 | 91.51 |
| **Average (English)** | | 74.23 | 73.96 | 79.07 | 76.72 | 77.45 | 74.16 | 77.60 | 77.92 |
| | OpenBookQA_sea | 88.98 | 83.49 | 91.90 | 90.02 | 83.29 | 79.74 | 84.07 | 89.94 |
| | XCOPA_sea | 98.51 | 97.12 | 98.51 | 96.55 | 98.72 | 94.58 | 98.63 | 97.64 |
| | MMLU_sea | 85.20 | 84.61 | 88.73 | 83.10 | 84.96 | 80.48 | 86.55 | 81.49 |
| | HellaSwag_sea | 75.43 | 78.39 | 86.34 | 85.38 | 83.97 | 80.67 | 83.80 | 86.92 |
| **Average (Multilingual)** | | 86.77 | 85.90 | 91.37 | 88.76 | 87.74 | 83.87 | 88.26 | 89.00 |

**Compass-v3 has demonstrated reasonably good performance in the general English field, performing comparably to proprietary models such as GPT-4.1-mini and GPT-4o-mini. In multilingual evaluation, Compass-v3 closely trails GPT-4o, with a performance difference of only 0.97%.** As shown in Table 6, Compass-v3 performs competitively across all evaluated domains, often ranking among the top performers in reading comprehension, common sense reasoning, and multilingual understanding. In particular, it achieves 89.79% on BoolQ and 90.89% on Belebele, indicating strong language comprehension capabilities. Its common sense reasoning abilities are particularly noteworthy, with scores of 86.92% on HellaSwag, 87.08% on Winogender, and 99.80% on XCOPA—ranking closely with the highest-performing models. In knowledge-based QA, Compass-v3 achieves 81.37% on MMLU, competitive with top-tier proprietary systems. Although its performance on ARC (61.86%) and GPQA (50.00%) trails behind the leading models, it remains within a reasonable range for a model of its size. Notably, Compass-v3 attains 91.28% on GSM8K, reflecting strong mathematical reasoning abilities and placing it among the top-performing models in this domain.

In the multilingual setting, Compass-v3 demonstrates strong cross-lingual generalization, achieving an average score of 86.77%, which is within 1 percentage point of GPT-4o. It performs particularly well on OpenBookQA_sea (88.98%) and XCOPA_sea (98.51%), showing minimal degradation relative to their English counterparts. Taken together, these results indicate that Compass-v3 offers a well-rounded performance profile across both English and multilingual tasks.

## 7. Conclusion

In this work, we presented Compass-v3, a vertical-domain Mixture-of-Experts (MoE) language model with 245B total parameters (71B active), designed to address the unique challenges of Southeast Asian e-commerce. By adopting fewer but larger experts, together with hardware-aware optimizations such as intra-node expert parallelism, memcpy operator improvements, and expert-aware FP8 quantization, Compass-v3 maximizes efficiency under limited compute resources. Trained on 12T tokens of curated multilingual corpora and synthetic e-commerce instructions, and aligned through Optimal-Transport Direct Preference Optimization (OTPO), the model achieves state-of-the-art results on in-house and open-source e-commerce benchmarks. It further demonstrates robust multilingual competence across low-resource Southeast Asian languages while preserving strong general performance, underscoring its dual strengths of specialized e-commerce expertise and broad linguistic competence.

Looking ahead, we plan to extend Compass-v3 in several directions: incorporating multimodal signals such as product images and videos to better capture real-world commerce contexts; developing more efficient inference methods, including adaptive expert routing and retrieval-augmented generation, to reduce deployment costs; expanding alignment with human-in-the-loop preference optimization across diverse cultural and linguistic settings; and enabling continual training pipelines to adapt to rapidly evolving products and user behaviors. We hope Compass-v3 provides both a practical system for multilingual commerce and a foundation for future research on domain-specialized LLMs.